\documentclass[journal]{IEEEtran}
\usepackage{amsmath,amsfonts}
\usepackage{algorithmic}
\usepackage{algorithm}
\usepackage{array}
\usepackage[caption=false,font=normalsize,labelfont=sf,textfont=sf]{subfig}
\usepackage{textcomp}
\usepackage{stfloats}
\usepackage{url}
\usepackage{verbatim}
\usepackage{graphicx}
\usepackage{cite}

\usepackage[utf8]{inputenc} 
\usepackage{nicefrac}       
\usepackage{multirow}
\usepackage{arydshln}
\usepackage{pifont}
\usepackage{tabularx}
\usepackage{booktabs}
\usepackage{threeparttable}
\usepackage{eso-pic}

\newcommand{\approach}{\text{FlexTSF}}
\newcommand{\cmark}{\ding{51}}%
\newcommand{\xmark}{\ding{55}}%

\graphicspath{ {./figures/} }

\begin{document}

\AddToShipoutPictureFG*{\AtPageLowerLeft{\put(\LenToUnit{\dimexpr 1in+\hoffset+\oddsidemargin\relax},\LenToUnit{0.2in}){\parbox[b]{\textwidth}{\scriptsize \copyright~2026 IEEE. Personal use of this material is permitted. Permission from IEEE must be obtained for all other uses, in any current or future media, including reprinting/republishing this material for advertising or promotional purposes, creating new collective works, for resale or redistribution to servers or lists, or reuse of any copyrighted component of this work in other works. DOI: 10.1109/TKDE.2026.3732095.}}}}

\title{FlexTSF: A Flexible Forecasting Model for Time Series with Variable Regularities}

\author{Jingge~Xiao, Yile~Chen, Gao~Cong, Wolfgang~Nejdl, and Simon~Gottschalk%
    \thanks{This work was supported in part by the German Federal Ministry for Transport (BMV) via mFUND, Germany (``MoToRes'') under Grant 01F2271A and in part by German Federal Ministry for Economic Affairs and Energy (BMWE), Germany (``ATTENTION!'') under Grant 01MJ22012D.}%
	\thanks{Jingge Xiao, Wolfgang Nejdl, and Simon Gottschalk are with the L3S Research Center, Leibniz University Hannover, Germany (emails: xiao@L3S.de, nejdl@L3S.de, gottschalk@L3S.de).}%
	\thanks{Yile Chen is with the School of Computer Science and Engineering, South China University of Technology, China (emails: chenyile@scut.edu.cn).}
	\thanks{Gao Cong is with College of Computing and Data Science, Nanyang Technological University, Singapore (emails: gaocong@ntu.edu.sg).}}



\maketitle

\begin{abstract}
	Generalizing across time series with irregular temporal structures poses a persistent challenge for universal time series forecasting models. Existing approaches often assume regular sampling or depend heavily on imputation, which limits their applicability in real-world scenarios where irregularities are prevalent due to diverse sensing devices and recording practices. To address this gap, we propose FlexTSF, a flexible and domain-adaptive forecasting model specifically designed for time series with variable temporal regularities. At its foundation lies the IVP Patcher, a continuous-time patching module leveraging Initial Value Problems (IVPs) to inherently handle uneven sampling intervals, variable sequence lengths, and missing values. Built upon a decoder-only Transformer architecture, FlexTSF integrates normalized timestamp inputs and domain-specific statistical descriptors through a specialized causal self-attention mechanism, enabling adaptability across domains. Extensive experiments on 16 datasets demonstrate FlexTSF’s effectiveness, significantly outperforming existing models in classic forecasting scenarios, zero-shot generalization, and low-resource fine-tuning conditions.
\end{abstract}


\section{Introduction}

Time series forecasting, the task of predicting future values based on historical observations, plays an indispensable role across numerous domains, including finance, retail, healthcare, and meteorology \cite{lim2021time,de200625tsf}. Apart from conventional supervised forecasting models~\cite{PatchTST:nie2023time, DLinear:zeng2023transformers, itransformer:liu2023itransformer}, significant research has been recently conducted toward developing universal time series forecasting models that are pre-trained once and directly applied to various domains \cite{MOIRAI:woo2024unified,lagllama:rasul2023lag, Chronos:ansari2024chronos}.  However, alongside these advances and the broader applicability, existing pre-trained models continue to face a fundamental challenge, namely \textbf{temporal irregularity}, which is characterized by the presence of missing values, uneven sampling intervals, and variable sequence lengths.

\begin{figure}[!t]
	\centering
	\includegraphics[width=1.00\linewidth]{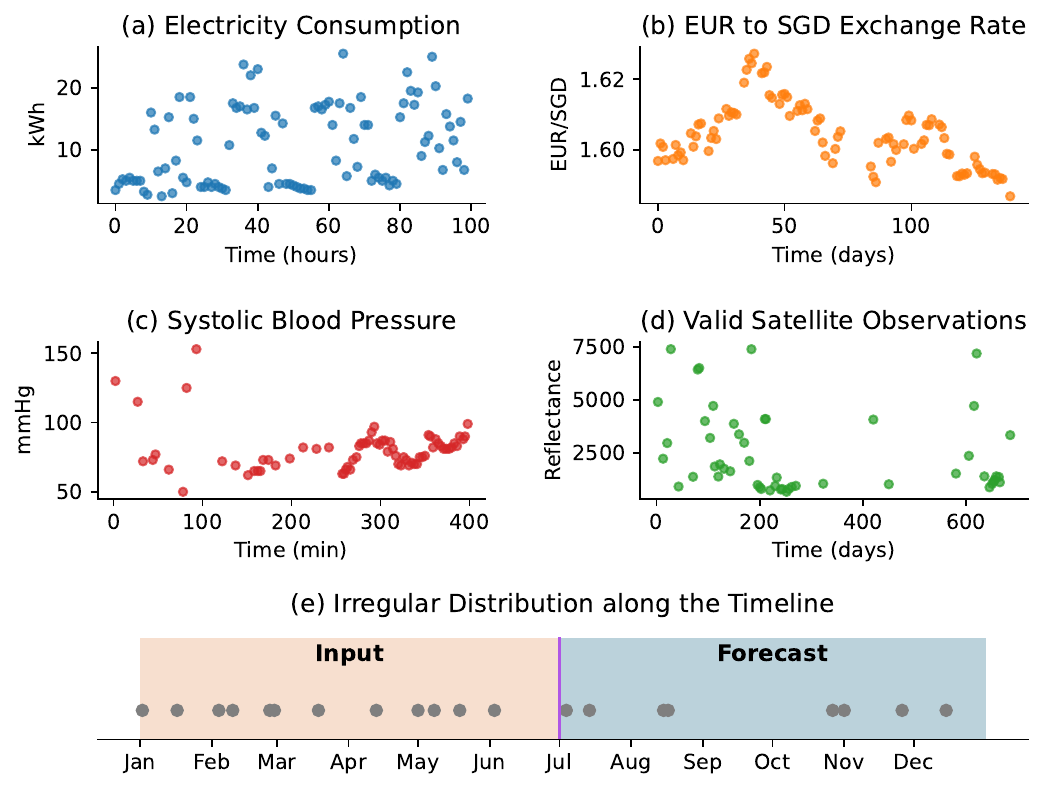}
	\caption{Illustrations of temporal irregularity.}
	\label{fig:diff_ts}
\end{figure}

We illustrate temporal irregularity in Fig.~\ref{fig:diff_ts}. (a) shows regularly sampled data; (b) depicts missing data due to events like holidays; (c) presents blood pressure measurements becoming denser as a patient's condition worsens; (d) shows irregular satellite observations influenced by atmospheric conditions such as clouds and fog; and (e) illustrates irregular temporal distributions. The temporal irregularity exists not only in input data. From the output perspective, users sometimes need forecasts only around specific time points. For instance, diabetic patients typically focus on glucose levels before and after meals, rather than tracking them for all 1,440 minutes of a day. Weather forecasts may alternate between long-term daily predictions and imminent rainfall probabilities.

However, existing universal time series forecasting models are developed under the setting of regular sampling~\cite{MOIRAI:woo2024unified, Moirai-MoE:liu2025, Chronos:ansari2024chronos}. Consequently, they are constrained to operate at a fixed temporal regularity for both inputs and predictions. When tackling temporal irregularity, these models require additional data augmentation (e.g., imputation) to align input sequences, and are compelled to operate at the finest granularity and produce redundantly long prediction sequences (e.g., 14,400 one-minute steps over ten days), leading to ineffectiveness and inefficiency in both computation and modeling.
Moreover, in large-scale pre-training and downstream applications, data from different domains (electricity consumption, exchange rates, systolic blood pressure, etc.) are naturally recorded at varying temporal granularities (e.g., minutes, hours, days), which further complicates universal generalization for a single model. These limitations make developing a model solution both essential and non-trivial.

To address these challenges, we propose \approach{}\footnote{Source code is at \url{https://github.com/jingge326/FlexTSF}.}, a flexible and domain-adaptive time series forecasting model that is capable of tackling temporal irregularity and accommodating varying granularities across domains. Built upon the widely adopted patch-based Transformer architecture for time series forecasting, \approach{} introduces novel modules to effectively organize the data with temporal irregularity and varying granularity into calibrated representations, and enable forecasting at arbitrary timestamps.

Specifically, \approach{} integrates IVP Patcher, a novel continuous-time, patch-level modeling module that captures temporal dynamics under temporal irregularity within patches. Unlike prior approaches that use fixed-size patches designed for regularly sampled series~\cite{PatchTST:nie2023time, MOIRAI:woo2024unified, TimesFM:das2024decoder}, IVP Patcher leverages Initial Value Problems (IVPs)~\cite{Bilos2021:neural-flow} to learn patch representations through differential equations over arbitrary sampling intervals. Moreover, IVP Patcher is equipped with strategies to adaptively determine patch lengths based on data characteristics.  To unify heterogeneous temporal granularities, \approach{} normalizes timestamps from diverse domains into a domain-invariant temporal unit, eliminating the need for manually engineered time features (e.g., year, hour, or second) and providing a consistent temporal scale across large pre-training corpora.

Based on the IVP Patcher for patch-level modeling, \approach{} extends its capacity to capture long-range temporal dependencies through causal self-attention within a Transformer backbone. To enhance domain adaptability, \approach{} conditions its generative process on domain-specific statistical descriptors and normalized temporal unit embeddings derived during the timestamp normalization stage. Furthermore, we design a unified training and forecasting pipeline grounded in probabilistic and variational inference, specifically tailored for irregularly sampled time series. This framework enables flexible and consistent forecasting at arbitrary target timestamps, effectively bridging irregular temporal dynamics with continuous-time prediction capabilities.

Our evaluation on 16 benchmark datasets demonstrates that \approach{} achieves the lowest MSE in 22 of 24 irregular-forecasting tasks (8 datasets × 3 horizons), exhibits better robustness across varying missing rates, and significantly outperforms state-of-the-art baselines in the zero-shot setting. Further analyses confirm the contribution of each component, the benefits of using random patch lengths, and the efficiency of the overall design.

We summarize our main contributions as follows:
\begin{itemize}
	\item We propose \approach{}, to the best of our knowledge, the first universal forecasting model built from the perspective of breaking data regularity constraints.
	\item We introduce IVP Patcher, a continuous-time patching module, to handle irregular time series, overcoming limitations of traditional fixed-size patching methods.
	\item We propose a timestamp normalization scheme and a unified training and forecasting pipeline for variable temporal irregularities, enabling domain-aware adaptation and improving cross-domain generalization.
	\item We conduct extensive evaluations demonstrating \approach{}'s superior forecasting capability in classic, zero-shot, and low-resource fine-tuning scenarios, and we quantify the positive effects of its components, random patch lengths, and computational efficiency.
\end{itemize}

\section{Related Work}

\subsection{Transformers for Time Series Forecasting}
Time series forecasting has evolved from classical statistical methods such as ARIMA~\cite{arima:box1970time} and GARCH~\cite{bollerslev1986generalized} to deep learning models based on RNNs~\cite{salinas2020deepar}, CNNs~\cite{borovykh2018dilated}, and more recently, Transformers~\cite{Informer:zhou2021informer}. Early Transformer-based models for time series forecasting, including LogTrans~\cite{LogTran:Shiyang2019}, Informer~\cite{Informer:zhou2021informer}, Autoformer~\cite{Autoformer:Wu2021AutoformerDT}, FEDformer~\cite{zhou2022fedformer}, and Pyraformer~\cite{liu2021pyraformer}, introduced various architectural innovations to reduce the quadratic complexity of self-attention and improve long-horizon forecasting. These include mechanisms like convolutional or ProbSparse attention, trend-seasonality decomposition, and frequency-domain transforms. Beyond these, PatchTST~\cite{PatchTST:nie2023time} replaces tokenization at individual timestamps with fixed-length patches to capture local temporal patterns. Following it, Crossformer~\cite{Crossformer:zhang2023} employs a two-stage attention mechanism alternating between temporal and inter-variable dimensions to model multivariate interactions. Spatio-temporal forecasting studies further model structured dependencies across variates using graph-enhanced Transformer designs~\cite{PMRGCN:gao2024,PDFormer:jiang2023,GESTT:kong2025}. Meanwhile, iTransformer~\cite{itransformer:liu2023itransformer} treats variables as tokens and time as channels, focusing attention on cross-variable patterns and using feed-forward layers for temporal modeling. These works laid the empirical foundation for developing universal time series forecasting models based on Transformers.

\subsection{Universal Time Series Forecasting Models}
Recent research has increasingly focused on training universal forecasting models~\cite{MOMENT:goswami2024moment}, which extend the Transformer lineage to achieve broad zero-shot generalization across diverse datasets rather than dataset-specific optimization. In Table \ref{tab:compare_tsf_models}, we provide a technical comparison of six representative open-source models with \approach{}. ForecastPFN~\cite{Forecastpfn:dooley2024forecastpfn} and DAM~\cite{DAM:darlowdam} use pointwise time-value pairs as input tokens within encoder-only models, but differ in their forecasting approaches: the former relies on time queries while the latter utilizes basis function composition. Lag-Llama~\cite{lagllama:rasul2023lag} and TimesFM~\cite{TimesFM:das2024decoder} employ decoder-only models, adopting autoregressive prediction, which is similar to next token prediction in LLMs, for training on time series data. MOIRAI~\cite{MOIRAI:woo2024unified} and its successor MOIRAI-MoE~\cite{Moirai-MoE:liu2025} utilize time series patches with multiple variables and leverage an encoder-only model trained via masked patch prediction. Sundial~\cite{Sundial:liu2025} is a family of decoder-only Transformer models that are pre-trained on continuous-valued time series using a TimeFlow loss based on flow-matching to generate multiple probable predictions. Most of these models are designed for well-structured and regularly sampled time series~\cite{ye2024survey}. DAM~\cite{DAM:darlowdam} considered irregularity when building the model, but the model was not applied to forecasting such data. Besides these models, Timer-XL~\cite{TimerXL:liu2025timerxl} extends the decoder-only paradigm to long-context, unified forecasting through multivariate next-token prediction, while Time-MoE~\cite{shi2025timemoe} scales a decoder-only mixture-of-experts architecture to 2.4B parameters. TimeMixer and TimeMixer++~\cite{TimeMixer:wang2024timemixer,TimeMixer++:wang2025timemixer} further explore MLP-based multiscale mixing architectures for forecasting. FlexTSF stands out because of its native support for missing data, variable sequence lengths, and irregular sampling intervals, making it well-suited to handle the heterogeneous data structure issues often encountered in real-world scenarios.

\begin{table*}[htbp]
	\centering
	\caption{Comparison Between Universal Time Series Forecasting Models}
	\footnotesize
	\setlength{\tabcolsep}{4pt}
	\begin{threeparttable}
		\begin{tabular}{lccccccc}
			\toprule
			                            & \textbf{ForecastPFN} & \textbf{Lag-Llama} & \textbf{DAM}    & \textbf{TimesFM} & \textbf{MOIRAI-MoE} & \textbf{Sundial} & \textbf{FlexTSF (Ours)} \\
			\midrule
			\textbf{Tokenization}       & Point-wise TV        & Lag Feature        & Point-wise TV   & Patch            & Patch               & Patch            & IVP Patcher             \\
			\textbf{Architecture}       & Encoder-Only         & Decoder-Only       & Encoder-Only    & Decoder-Only     & Encoder-Only        & Decoder-Only     & Decoder-Only            \\
			\textbf{Forecast}           & Time Query           & Autoregression     & Basis Functions & Autoregression   & Masked Prediction   & Autoregression   & Direct Multi-Horizon    \\
			\textbf{Variate}            & Univariate           & Univariate         & Univariate      & Univariate       & Multi/Univariate    & Univariate       & Univariate              \\
			\textbf{Probabilistic}      & \xmark               & \cmark             & \xmark          & \xmark           & \cmark              & \cmark           & \cmark                  \\
			\textbf{Missing Values}     & \xmark               & \xmark             & \cmark          & \xmark           & \xmark              & \xmark           & \cmark                  \\
			\textbf{Various Length}     & \cmark               & \cmark             & \cmark          & \cmark           & \xmark              & \cmark           & \cmark                  \\
			\textbf{Irregular Interval} & \cmark               & \xmark             & \cmark          & \xmark           & \xmark              & \xmark           & \cmark                  \\
			\bottomrule
		\end{tabular}
		\label{tab:compare_tsf_models}

		\begin{tablenotes}[para] 
			\scriptsize
			\noindent
			"TV" indicates time-value pairs. \cmark (or \xmark) indicates the model (does not) naturally support this kind of data characteristic.
		\end{tablenotes}

	\end{threeparttable}
\end{table*}

\subsection{Irregular Time Series Modeling}
Irregular time series are becoming increasingly prevalent due to the widespread adoption of various sensing devices and recording practices~\cite{weerakody2021review}. In recent years, substantial progress has been made in developing models to handle irregular time series, as demonstrated by works such as GRU-D~\cite{GRUD:Che2018grud}, Latent-ODE~\cite{latentode:rubanova2019latent}, mTAN~\cite{mTAN:Shukla2021mTAN}, CRU~\cite{cru:schirmer2022modeling}, ContiFormer~\cite{Contiformer:chen2023}, and t-PatchGNN~\cite{t-PatchGNN:zhang2024irregular}. However, none of these works has succeeded in building and pre-training a universal time series forecasting model that is broadly applicable across diverse domains and regularities without additional post-training.
On the other hand, existing continuous latent variable models such as Latent-ODE~\cite{latentode:rubanova2019latent} and IVP-VAE~\cite{ivpvae:xiao2024ivp} treat the entire time series as a single sample for variational inference and posit a latent trajectory that evolves under neural ODEs or flows. Observations are decoded from this trajectory, and forecasts are obtained by extrapolating it. When the trajectory struggles to accommodate biases introduced by new observations, researchers incorporate state jumps~\cite{gob:DeBrouwer2019,neural_jump:jia2019neural}. A major drawback of this paradigm is that a single trajectory must carry all long-range structure, distributional shifts, and domain characteristics, forcing the decoder to generate forecasts from a bottlenecked representation that limits expressiveness and generalization. \approach{} instead factorizes the sequence into patches, learns a latent variable for each patch, and composes them causally via decoder-only attention to predict future patch latents, enabling long-horizon conditioning without stepwise trajectory rolling. Moreover, unlike imputation-based approaches that fill missing entries on a regular grid~\cite{BRITS:cao2018brits, CSDI:tashiro2021csdi}, \approach{} uses only available time-value observations as input. The IVP Patcher learns latent continuous-time dynamics directly from these observations. Irregularity is handled in latent patch space rather than through an imputed observation sequence.

\section{Problem Formulation \& Prerequisites}
\label{subsec:problem_formulation}
\paragraph{Problem Formulation}
We consider a time series $S$ as a sequence of $M$ elements $S=\{(\mathbf{x}_i, t_i)\}_{i=1}^M$, where each element $(\mathbf{x}_i, t_i)$ consists of a multivariate observation $\mathbf{x}_i$ collected at timestamp $t_i$. This formulation is flexible to accommodate a diverse range of time series characterized by different sequence lengths (varying $M$ across samples), irregular sampling intervals ($t_{i+1} - t_{i} \neq t_{i} - t_{i-1}$), and incomplete observations (missing variables in $\mathbf{x}_i$).
Referencing previous works~\cite{PatchTST:nie2023time, OneFitsAll:zhou2023one, DAM:darlowdam}, we address multivariate time series in a channel-independent manner.
Our goal is to develop a universal forecasting model that takes any time series $S$ as input and predicts its future values
$\{\mathbf{x}_i\}_{i=M+1}^{M+H}$ over a subsequent time window $\{t_i\}_{i=M+1}^{M+H}$ at arbitrary timestamps, where $H$ denotes the forecast horizon.
Key notations are summarized in Table~\ref{tab:notation}. A patch is precisely defined as \(V_k=\{(\mathbf{x}_{k,m}, t_{k,m})\}_{m=1}^{\ell}\)
with start time \(t_{k,0}\) and latent states \(\mathbf{z}_{k,m}\). To maintain readability, in the following descriptions,
when focusing on content within a single patch, we omit the patch index and write \(\mathbf{x}_m\), \(t_m\), \(t_{0}\), \(\mathbf{z}_m\), etc.
For patch-level operations (e.g., causal self-attention, ELBO), we keep the patch index and omit within-patch indices, using \(\mathbf{r}_k\), \(V_k\), and \(t_{k}\).

\begin{table}[htbp]
	\footnotesize
	\centering
	\caption{Summary of Notations}
	\begin{tabularx}{\linewidth}{lX}
		\toprule
		\textbf{Notation}                & \textbf{Description}                                              \\ \midrule
		$S$                              & A time series sample $\{(\mathbf{x}_i,t_i)\}_{i=1}^M$             \\
		$M$                              & Number of observations in $S$                                     \\
		$H$                              & Forecast horizon (number of future steps)                         \\
		$\mathcal{X}$                    & Observed value sequence $\{\mathbf{x}_i\}_{i=1}^M$                \\
		$T$                              & Observed timestamps $\{t_i\}_{i=1}^M$                             \\
		$L$                              & Static vector $[\mu_g,\sigma_g,\mu_s,\sigma_s,\omega_g,\omega_s]$ \\
		$V_k$                            & Patch $k$: $\{(\mathbf{x}_{k,m}, t_{k,m})\}_{m=1}^{\ell}$         \\
		$m$                              & Within-patch index $1,\ldots,\ell$                                \\
		$K_{\text{in}},K_{\text{out}},K$ & Number of observed, dummy, and total patches                      \\
		$\ell$                           & Patch length                                                      \\
		$t_{k,0}, t_{k}, t_{0}$          & Start time of patch $k$                                           \\
		$\mathbf{z}(t)$                  & Latent trajectory of IVP; $\mathbf{z}(t_{k,0})=\mathbf{r}_k$      \\
		$\mathbf{r}_k$                   & Patch-level latent variable; sample used as representation        \\
		$\hat{\mathbf{r}}_j$             & Predicted future patch latent for $j>K_{\text{in}}$               \\
		\bottomrule
	\end{tabularx}
	\label{tab:notation}
\end{table}

\paragraph{Patching}

Patching is a common strategy to obtain time series representations by organizing a sequence into fixed-length segments, facilitating efficient modeling with transformers~\cite{PatchTST:nie2023time, TimesFM:das2024decoder}. We denote patch $k$ by $V_k=\{(\mathbf{x}_{k,m}, t_{k,m})\}_{m=1}^{\ell}$. Each patch aggregates a consecutive segment of length $\ell$ and serves as a modeling unit. Classic patching methods use a predefined $\ell$ for each dataset with regular time intervals, often requiring careful tuning for optimal performance, and do not consider the time information.

\paragraph{Initial Value Problems}

An initial value problem (IVP)~\cite{sauer2012numerical} models the evolution of a system's state by solving an ordinary differential equation (ODE) of the form
\begin{equation}
	\frac{d\mathbf{z}(t)}{dt} = f(t, \mathbf{z}(t))\text{,}
	\label{eq:ivp}
\end{equation}
with initial condition $\mathbf{z}(t_0) = \mathbf{z}_0$, where $\mathbf{z}(t)$ denotes the latent state at time $t$, $f$ describes the system’s time-dependent dynamics, and $\mathbf{z}_0$ is the initial latent state at $t_0$.

Two characteristics of the IVP formulation in Eq.~\eqref{eq:ivp} make it suitable for patching irregular time series:
1) The derivative $f$ naturally accommodates data with irregular temporal distributions.
2) $f(t, \mathbf{z}(t))$ is capable of modeling time-dependent dynamics, which is ideal for time series that are often driven by internal or external factors varying over time (complex, non-stationary behaviors).
Pioneering works such as Neural ODEs~\cite{Chen2018:neural-ode} and Neural Flows~\cite{Bilos2021:neural-flow} provided fundamental tools for learning continuous-time, time-dependent dynamics from irregularly sampled time series.

\paragraph{Value Normalization}\label{sec:value_norm}

Recent research in time series forecasting often adopts a two-step normalization scheme for input values: global and series-level normalization. Given a value sequence \(\mathcal{X} = \{\mathbf{x}_i\}_{i=1}^M\) from dataset \(D\), global normalization $G_g$ computes the mean \(\mu_g\) and standard deviation \(\sigma_g\) over \(D\), then standardizes each value. Series-level normalization \(G_s\)~\cite{PatchTST:nie2023time} further standardizes each sequence using its own mean \(\mu_s\) and standard deviation \(\sigma_s\). Formally: $\operatorname{ValueNorm}(\mathcal{X}) := G_s(G_g(\mathcal{X}))$, yielding normalized values. While value normalization is extensively applied, normalization of temporal information remains largely overlooked.

\section{Methodology}

\subsection{Overview}

\begin{figure*}[t]
	\centering
	\includegraphics[width=0.9\linewidth]{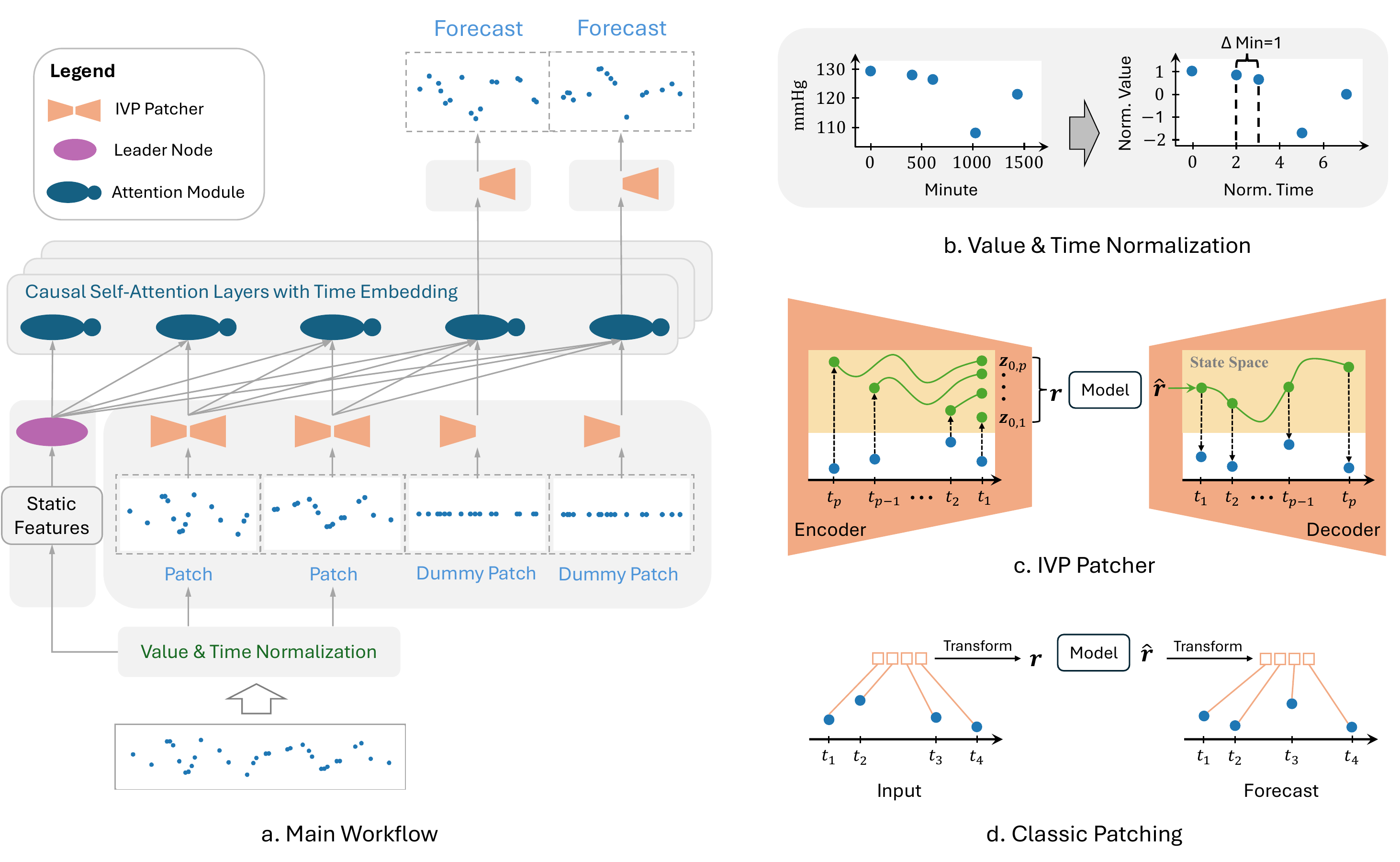}
	\caption{\textbf{Framework Overview.} (a) \approach{} normalizes input time series, extracts patch-level latent variables via the IVP Patcher, applies self-attention with a Leader node, and decodes dummy patches into forecasts. (b) A time series with irregular intervals is transformed into a numerical scale with the minimum interval equal to 1. (c) IVP Patcher models continuous-time dynamics on irregular data by solving IVPs. (d) Classic patching directly transforms fixed-size sequences while ignoring time information.}
	\label{fig:FlexTSF}
\end{figure*}

An overview of \approach{}'s dataflow is illustrated in Fig.~\ref{fig:FlexTSF}(a). \approach{} begins by applying value and time normalization to the raw input sequence, bringing observations into a unified numeric and temporal scale. The normalized series is then split into patches, with dummy patches appended at the end to represent the forecast horizon. Each patch $V_k$ is processed by the encoder of the IVP Patcher to infer a patch-level latent variable $\mathbf{r}_k$ with variational posterior $q(\mathbf{r}_k\mid V_k)$, from which we draw samples as the patch representation. This latent variable is used in two ways: it is decoded by the IVP Patcher to reconstruct the observed patch, and it is fed to a causal self-attention stack. Unlike continuous-time latent models~\cite{latentode:rubanova2019latent, ivpvae:xiao2024ivp} that extrapolate time series by rolling a single latent trajectory forward, we factorize the generative process at the patch level and causally predict future patch latents $\hat{\mathbf{r}}$ from past patch latents, which then initialize the decoder part for IVP Patcher to produce forecasting results.

The self-attention stack processes the sequence of patch latents, along with static features provided by a Leader node and time embeddings applied at each attention layer. After the attention stack, hidden states of dummy patches are used as forecast anchors, which are passed to the decoder part of the IVP Patcher to generate forecasts by evolving forward to target timestamps. The model is trained end-to-end using a combination of patch-level ELBO and forecast log-likelihood losses, with the IVP Patcher focused on learning local dynamics and the attention stack capturing higher-level temporal dependencies across patch latents. Algorithm~\ref{alg:flexTSF} provides the step-by-step pseudocode for the proposed framework. Next, we present the details of each component.

\begin{algorithm}[htbp]
	\caption{FlexTSF for Forecast.}
	\label{alg:flexTSF}
	{\raggedright \textbf{Inputs}: Observed values $\mathcal{X}$, timestamps $T$ and future timestamps $\{t_i\}_{i=M+1}^{M+H}$ \\}
	{\raggedright \textbf{Outputs}: Predicted future values $\{\hat{\mathbf{x}}_i\}_{i=M+1}^{M+H}$\\}
	\begin{algorithmic}[1]
		\STATE Apply $\operatorname{ValueNorm}(\mathcal{X})$ and $\operatorname{TimeNorm}(T, \{t_i\}_{i=M+1}^{M+H})$
		\STATE $\{V_k\}_{k=1}^{K_{\text{in}}} \leftarrow \operatorname{Partition}((\mathcal{X},T),\ell)$
		\STATE Append $K_{\text{out}}$ dummy patches; set $K = K_{\text{in}} + K_{\text{out}}$
		\STATE For all patches ($k = 1,\ldots,K$), each patch has
		\STATE \hspace{\algorithmicindent} $\{\mathbf z_m\}_{m=1}^{\ell}\leftarrow\operatorname{Linear}(\{\mathbf x_{m}\}_{m=1}^{\ell})$
		\STATE \hspace{\algorithmicindent} $\{\mathbf z_{0,m}\}_{m=1}^{\ell}\leftarrow\{\operatorname{EvolveState}(\mathbf z_m, t_{m}, t_{0})\}_{m=1}^{\ell}$
		\STATE \hspace{\algorithmicindent} $q(\mathbf r_k\mid V_k)\leftarrow\operatorname{PosteriorMixture}(\{\mathbf z_{0,m}\}_{m=1}^{\ell})$
		\STATE \hspace{\algorithmicindent} Sample $\mathbf r_k\sim q(\cdot\mid V_k)$

		\STATE Each observed patch ($k = 1,\ldots,K_{\text{in}}$) has
		\STATE \hspace{\algorithmicindent} $\mathbf z_0\leftarrow\mathbf r_k$
		\STATE \hspace{\algorithmicindent} $\{\mathbf z_m\}_{m=1}^{\ell}\leftarrow\{\operatorname{EvolveState}(\mathbf z_0, t_{0}, t_{m})\}_{m=1}^{\ell}$
		\STATE \hspace{\algorithmicindent} $\{\hat{\mathbf x}_{m}\}_{m=1}^{\ell}\leftarrow\operatorname{Linear}(\{\mathbf z_m\}_{m=1}^{\ell})$

		\STATE Forecast: $\{\hat{\mathbf r}_j\}_{j=K_{\text{in}}+1}^{K} \leftarrow f_\theta ([L,\, \{\mathbf{r}_k\}_{k=1}^{K_{\text{in}}}], \{t_{k,0}\}_{k=1}^{K} )$

		\STATE Each forecast patch ($j = K_{\text{in}}+1,\ldots,K$) has
		\STATE \hspace{\algorithmicindent} $\mathbf z_0\leftarrow\hat{\mathbf r}_{j}$
		\STATE \hspace{\algorithmicindent} $\{\mathbf z_m\}_{m=1}^{\ell}\leftarrow\{\operatorname{EvolveState}(\mathbf z_0, t_{0}, t_{m})\}_{m=1}^{\ell}$
		\STATE \hspace{\algorithmicindent} $\{\hat{\mathbf x}_{m}\}_{m=1}^{\ell}\leftarrow\operatorname{Linear}(\{\mathbf z_m\}_{m=1}^{\ell})$
		\STATE Concatenate $K_{\text{out}}$ forecast patches to form $\{\hat{\mathbf{x}}_i\}_{i=M+1}^{M+H}$
		\STATE Return $\{\hat{\mathbf{x}}_i\}_{i=M+1}^{M+H}$.
	\end{algorithmic}
\end{algorithm}

\subsection{Conditioned Latents for Forecasting}\label{sec:con_gen_forecast}

We formulate \approach{} as a generative model. Given the joint distribution over all patches $p(\{V_k\}_{k=1}^{K})$, the observed and future patches can be separated via
\begin{equation}
	\begin{aligned}
		\log p(\{V_k\}_{k=1}^{K})
		= & \log p(\{V_k\}_{k=1}^{K_{\text{in}}})                                       \\
		  & + \log p(\{V_k\}_{k=K_{\text{in}}+1}^{K}\mid\{V_k\}_{k=1}^{K_{\text{in}}}).
	\end{aligned}
	\label{eq:joint_in_out}
\end{equation}

In a generic latent variable model, latent variables $\mathbf{r}$ are introduced together with a prior $p(\mathbf{r})$ and a likelihood $p(V\mid\mathbf{r})$, yielding a joint distribution $p(V,\mathbf{r}) = p(\mathbf{r})\,p(V\mid\mathbf{r})$ and a marginal $p(V) = \int p(V,\mathbf{r})\,d\mathbf{r}$. Thus, regarding $p(\{V_k\}_{k=1}^{K_{\text{in}}})$, we introduce latent variables $\mathbf{r}$ with
\begin{equation}
	p(\{V_k\}_{k=1}^{K_{\text{in}}})=\int p(\{V_k\}_{k=1}^{K_{\text{in}}}, \{\mathbf{r}_k\}_{k=1}^{K_{\text{in}}})\, d\{\mathbf{r}_k\}_{k=1}^{K_{\text{in}}}.
	\label{eq:marginal-latents}
\end{equation}
To obtain a tractable factorization of Eq.~\eqref{eq:marginal-latents}, FlexTSF assumes independent patch-latent priors $p(\mathbf{r}_k)=\mathcal{N}(\mathbf{0},\mathbf{I})$, so the joint distribution over observed patches factorizes as
\begin{equation}
	\begin{aligned}
		p(\{V_k\}_{k=1}^{K_{\text{in}}},\{\mathbf{r}_k\}_{k=1}^{K_{\text{in}}})
		 & = \prod_{k=1}^{K_{\text{in}}} p(\mathbf{r}_k)\, p\big(V_k \mid \mathbf{r}_k\big).
	\end{aligned}
	\label{eq:joint-contextual}
\end{equation}
Thus, for latent variable inference, we have:
\begin{equation}
	q(\{\mathbf{r}_k\}_{k=1}^{K_{\text{in}}}\mid \{V_k\}_{k=1}^{K_{\text{in}}})
	= \prod_{k=1}^{K_{\text{in}}} q(\mathbf{r}_k \mid V_k).
	\label{eq:q-factorized}
\end{equation}
With the joint factorization in Eq.~\eqref{eq:joint-contextual} and ~\eqref{eq:q-factorized}, this yields the sequence-level evidence lower bound (ELBO) over observed patches:
\begin{equation}
	\begin{aligned}
		\log p(\{V_k\}_{k=1}^{K_{\text{in}}})
		 & \ge \sum_{k=1}^{K_{\text{in}}} \mathbb{E}_{q(\mathbf{r}_k\mid V_k)}\big[\log p(V_k \mid \mathbf{r}_k)\big] \\
		 & \quad - \sum_{k=1}^{K_{\text{in}}} \mathrm{KL}\big(q(\mathbf{r}_k\mid V_k) \,\|\, p(\mathbf{r}_k)\big).
	\end{aligned}
	\label{eq:elbo-general}
\end{equation}

We define the patch-level loss using the right side of Eq.~\eqref{eq:elbo-general} as $\mathcal{L}_{\text{patch}}$, which is agnostic to the specific parameterization of $q(\mathbf{r}_k\mid V_k)$ and $p(V_k\mid\mathbf{r}_k)$. In Section~\ref{sec:ivp_patcher}, we instantiate this patch-level latent model with an IVP-based encoder--decoder, termed the IVP Patcher, and derive the corresponding implementation of $\mathcal{L}_{\text{patch}}$.

We now describe how these patch latents are conditioned for forecasting. Regarding $p(\{V_k\}_{k=K_{\text{in}}+1}^{K}\mid\{V_k\}_{k=1}^{K_{\text{in}}})$, we introduce predicted future patch latents $\{\hat{\mathbf{r}}_j\}_{j=K_{\text{in}}+1}^{K}$ conditioned on observed patch latents $\{\mathbf{r}_k\}_{k=1}^{K_{\text{in}}}$. We parameterize the conditional distribution with a Transformer-based causal self-attention network $f_\theta$:
\begin{equation}
	\{\hat{\mathbf{r}}_j\}_{j=K_{\text{in}}+1}^{K} = f_\theta ([L,\, \{\mathbf{r}_k\}_{k=1}^{K_{\text{in}}}], \{t_{k}\}_{k=1}^{K} ).
	\label{eq:cond-forecast}
\end{equation}
In Eq.~\eqref{eq:cond-forecast}, $L$ denotes the feature statistics of the data (details in Section~\ref{sec:timenorm_domainadapt}). Departing from the autoregressive chain rule, we assume that future latents are conditionally independent of each other given the observed patch latents and their own timestamps:
\begin{equation}
	\begin{aligned}
		 & p\bigl(\{\hat{\mathbf r}_j\}_{j=K_{\text{in}}+1}^{K} \mid \{\mathbf{r}_k\}_{k=1}^{K_{\text{in}}}, \{t_{k}\}_{k=1}^{K}\bigr) \\
		 & \approx \prod_{j=K_{\text{in}}+1}^K p\bigl(\hat{\mathbf r}_j\mid \{\mathbf{r}_k\}_{k=1}^{K_{\text{in}}},\,t_{j}\bigr).
	\end{aligned}
	\label{eq:conditioned_latents}
\end{equation}
This formulation removes teacher forcing, prevents error accumulation, and enables a single forward pass for all horizons.
The localized factorization and causal self-attention stack preserve a coherent conditioning pathway from inputs to forecasts while matching test-time causality with no future leakage. Because of this conditional independence, all $K_{\text{out}}$ future latents can be decoded in parallel: we pad the forecasting horizon with $K_{\text{out}}$ dummy patches (Algorithm~\ref{alg:flexTSF}, line~3) and read off the predicted latents $\{\hat{\mathbf{r}}_j\}_{j=K_{\text{in}}+1}^{K}$ at those positions in a single forward pass of the causal self-attention stack (Algorithm~\ref{alg:flexTSF}, line~13). Using dummy patches frames the forecasting process as predicting latent variables at future patch positions, keeping the generative process consistent across observed and future patches.

Although Eq.~\eqref{eq:conditioned_latents} removes the explicit autoregressive chain among future latents, it does not ignore future-to-future temporal dynamics. The conditional-independence assumption applies at the \emph{patch-latent} level, not at the observation level: within each patch, the IVP Patcher decoder (Section~\ref{sec:ivp_patcher}) propagates a single initial latent through continuous time, so consecutive forecasted observations within a patch are tightly coupled by the underlying dynamics. Across patches, all predicted latents $\{\hat{\mathbf{r}}_j\}$ are conditioned on the same shared context and receive layer-wise timestamp embeddings (Section~\ref{sec:timenorm_domainadapt}), which implicitly align adjacent future latents without an explicit autoregressive coupling.

Given a time series, the exact marginal likelihood implied by Eq.~\eqref{eq:conditioned_latents} is intractable. Instead, we map predicted latents to Gaussian distributions over future patches via the patch decoder $p(V_k\mid\mathbf{r}_k)$ (instantiated by the IVP Patcher in Section~\ref{sec:ivp_patcher}) and obtain the forecasting objective $\mathcal{L}_{\text{forecast}}$ through the likelihood of forecasted distributions against ground truth:
\begin{equation}
	\mathcal{L}_{\text{forecast}}
	= -\sum_{h=1}^{H}
	\log\mathcal{N}\bigl(
	\mathbf{x}_{M+h}\mid\hat{\mathbf{x}}_{M+h},
	\sigma^{2}\mathbf{I}
	\bigr),
	\label{eq:forecast-loss}
\end{equation}
where $\hat{\mathbf{x}}_{M+h}$ is decoded from $\hat{\mathbf{r}}_{j}$ of corresponding dummy patches. We adopt a fixed variance $\sigma^{2}$ following prior work~\cite{Chen2018:neural-ode,Bilos2021:neural-flow}.

Echoing Eq.~\eqref{eq:joint_in_out}, \approach{} is trained end-to-end by optimizing the joint objective:
\begin{equation}
	\mathcal{L} =\lambda \mathcal{L}_{\text{patch}} + \mathcal{L}_{\text{forecast}},
	\label{eq:joint-loss}
\end{equation}
Eq.~\eqref{eq:joint-loss} unifies the ELBO of observed time series $\mathcal{L}_{\text{patch}}$ (defined later in Eq.~\eqref{eq:patch-loss}) with the forecast likelihood $\mathcal{L}_{\text{forecast}}$ in Eq.~\eqref{eq:forecast-loss}, with $\lambda$ balancing the two objectives. The model can be trained in both supervised and self-supervised settings, where in self-supervised mode, the model learns to forecast hidden segments within the observed sequence. For time series with irregularity, the model calculates the loss at all available observation times. The training-time dataflow is as follows:
\begin{enumerate}
	\item Encode each observed patch $k$ with the IVP Patcher to obtain $\mathbf{r}_k\sim q(\mathbf{r}_k\mid V_k)$.
	\item Compute $\{\hat{\mathbf{r}}_j\}_{j=K_{\text{in}}+1}^{K}=f_\theta ([L,\, \{\mathbf{r}_k\}_{k=1}^{K_{\text{in}}}], \{t_{k}\}_{k=1}^{K} )$ with causal self-attention.
	\item Decode observed patches from $\mathbf{r}_k$ and decode dummy patches from $\hat{\mathbf{r}}_{j}$.
\end{enumerate}

\subsection{IVP Patcher}\label{sec:ivp_patcher}

The IVP Patcher is a module designed to handle temporal irregularity by replacing fixed-size patching with a continuous-time formulation. As shown in Fig.~\ref{fig:FlexTSF}(c), it models time series in each patch as indirect observations (blue dots) of underlying continuous latent processes (green lines) governed by ODEs. This contrasts with classic patching (Fig.~\ref{fig:FlexTSF}(d)), which extracts and transforms fixed-size value sequences while ignoring time information.

\subsubsection{Latent Inference} Given a patch $k$, a linear layer maps each observed data point $(\mathbf{x}_{m}, t_{m})$ to an initial latent state $\mathbf{z}_{m}$. To capture the temporal dynamics within the patch and establish a unified representation, we employ an efficient IVP solver~\cite{Bilos2021:neural-flow} to evolve these latent states backward in time to the starting timestamp of the patch $t_{0}$. Specifically, for each data point $(\mathbf{x}_{m}, t_{m})$, the corresponding latent state $\mathbf{z}_{m}$ serves as the initial condition at time $t_{m}$. We compute
\begin{equation*}
	\{\mathbf z_{0,m}\}_{m=1}^{\ell}=\{\operatorname{EvolveState}(\mathbf z_m, t_{m}, t_{0})\}_{m=1}^{\ell}
\end{equation*}
where $\operatorname{EvolveState}$ denotes the IVP solver that computes the aligned latent state $\mathbf{z}_{0,m}$ at reference time $t_{0}$ by integrating backward over the time difference $\Delta t_{m} = t_{0} - t_{m}$. This backward evolution aggregates information from different time points to a common temporal reference. It is applied simultaneously to all data points within the patch.

As discussed in Section~\ref{sec:con_gen_forecast}, we define the patch-level latent variable $\mathbf{r}_k$ to learn a latent distribution that encapsulates the essential information within the patch. Given evolved latent states $\{\mathbf{z}_{0,m}\}_{m=1}^{\ell}$, we infer $q_{\phi}(\mathbf{r}_k \mid V_k)$ to approximate the intractable true posterior in the VAE framework. We denote by $q_{\phi}$ and $p_{\theta}$ the variational posterior and likelihood parameterized by neural networks. The $\operatorname{PosteriorMixture}$ function (Algorithm~\ref{alg:flexTSF}) models this posterior as a mixture of diagonal Gaussian distributions:
\begin{equation}\label{eq:mixture}
	q_{\phi}(\mathbf{r}_k \mid V_k) = \sum_{m=1}^{\ell} \pi_m \, \mathcal{N}\big(\mathbf{r}_k \mid \boldsymbol{\mu}_{\mathbf{z}_{0,m}}, \, \boldsymbol{\sigma}_{\mathbf{z}_{0,m}}\big),
\end{equation}
where $\{\pi_m\}_{m=1}^{\ell}$ are the mixing coefficients for $\ell$ components, with weights given by
\begin{equation}
	\pi_m = \frac{1/\text{KL}_m}{\sum_j 1/\text{KL}_j + \epsilon}\text{,}
	\label{eq:pi-m}
\end{equation}
where $ \text{KL}_m = \text{KL}[\mathcal{N}(\boldsymbol{\mu}_{\mathbf{z}_{0,m}}, \boldsymbol{\sigma}_{\mathbf{z}_{0,m}}^2) \| \mathcal{N}(\mathbf{0}, \mathbf{I})] $.

To sample $\mathbf{r}_k$ from the mixture in Eq.~\eqref{eq:mixture}, we independently draw $ \tilde{\mathbf{z}}_{0,m} \sim \mathcal{N}(\boldsymbol{\mu}_{\mathbf{z}_{0,m}}, \boldsymbol{\sigma}_{\mathbf{z}_{0,m}}^2) $ and compute a KL-weighted average using $\pi_m$ from Eq.~\eqref{eq:pi-m}: $\mathbf{r}_k = \sum_{m=1}^\ell \pi_m \, \tilde{\mathbf{z}}_{0,m}$. This approximates the mixture's mean while emphasizing higher‑information components, and therefore yields a more concentrated, low-variance latent (Eq.~\eqref{eq:rk-dist}):
\begin{equation}
	\mathbf{r}_k \sim \mathcal{N}\!\left(\sum_m \pi_m \, \boldsymbol{\mu}_{\mathbf{z}_{0,m}},\; \sum_m \pi_m^2 \, \boldsymbol{\sigma}_{\mathbf{z}_{0,m}}^2\right)\textbf{,}
	\label{eq:rk-dist}
\end{equation}
where each component's mean $\boldsymbol{\mu}_{\mathbf{z}_{0,m}}$ and standard deviation $\boldsymbol{\sigma}_{\mathbf{z}_{0,m}}$ are parameterized by neural networks taking $\mathbf{z}_{0,m}$ as input.

\subsubsection{Patch Generation} \label{sec:forward_ivp}
As shown in Algorithm~\ref{alg:flexTSF}, the patch latents are used across time to reconstruct observed patches (from $\mathbf{r}_k$) and produce forecasts (from $\hat{\mathbf{r}}_j$). Both the reconstruction and forecasting processes utilize the same IVP Patcher decoder:
\begin{equation*}
	\{\mathbf z_m\}_{m=1}^{\ell}=\{\operatorname{EvolveState}(\mathbf z_0, t_{0}, t_{m})\}_{m=1}^{\ell}
\end{equation*}
where the initial state $\mathbf{z_0}$, initial time $t_{0}$, and target times $\{t_{m}\}_{m=1}^{\ell}$ vary depending on the position (observed or dummy) of the patch.

At the sequence level, dummy patch latents are conditioned on past observed ones via causal self-attention $f_\theta$, producing predicted latents at output positions $\hat{\mathbf{r}}_{j}$ for $j=K_{\text{in}}+1,\ldots,K$. Each latent initializes the decoder at the start time of the patch, i.e., $\mathbf{z}_{0} = \hat{\mathbf{r}}_j$. We then employ the IVP Patcher to evolve this latent state forward in time to the desired timestamps within the patch $\{t_{m}\}_{m=1}^{\ell}$. For each timestamp $t_{m}$, the IVP solver computes the corresponding latent state $\mathbf{z}_{m}$ by integrating forward over the time difference $\Delta t_{m} = t_{m} - t_{0}$. The evolution to all timestamps within a forecast patch is done simultaneously, as the computation of $\mathbf{z}_{m}$ at a specific time $t_{m}$ only depends on the initial state $\mathbf{z}_{0}$ which is the same within a patch. Finally, a linear layer maps these latent states to predicted values $\{\hat{\mathbf{x}}_{m}\}_{m=1}^{\ell}$.

These values parameterize an isotropic Gaussian likelihood for the patch,
\begin{equation}
	p_\theta\big(V_k \mid \mathbf{r}_k\big) \,=\, \prod_{m=1}^{\ell} \mathcal{N}\!\bigl(\mathbf x_{m} \mid \hat{\mathbf x}_{m},\;\sigma^2\mathbf I\bigr),
	\label{eq:patch-likelihood}
\end{equation}
which is used in the generation term of the patch loss. Combining Eq.~\eqref{eq:patch-likelihood} with the KL regularization, the patch-level loss is further detailed as:
\begin{equation}
	\mathcal{L}_{\text{patch}} = \sum_{k=1}^{K_{\text{in}}} (\frac{1}{\ell} \sum_{m=1}^\ell \text{KL}_m + \beta \cdot \mathbb{E}_{q_\phi} \left[ -\log p_\theta\big(V_k \mid \mathbf{r}_k\big) \right])
	\label{eq:patch-loss}
\end{equation}
where $\beta$ is a hyperparameter to balance the KL regularization and generation terms. Generation terms are computed on available entries when missing values exist.

Across all backward and forward solves in $\operatorname{EvolveState}$, we instantiate the IVP solver with ResNetFlow~\cite{Bilos2021:neural-flow}, a flow-based parameterization of the dynamics that avoids expensive adaptive ODE integration and is more efficient in practice than Neural ODE–style solvers while preserving continuous-time modeling capacity.

\subsubsection{Patch Length}\label{subsec:random_patches}

The IVP Patcher offers the advantage of dynamically selecting patch lengths for each input. We propose two strategies for generating patch lengths: deterministic and random. Compared to a single fixed patch length (a global hyperparameter used for every input), both strategies are adaptive.

In the deterministic approach, the patch length is chosen as a reproducible function of input characteristics (e.g., sequence length and forecast ratio), allowing inputs with different properties to be processed using corresponding patch lengths. This addresses the limitation of using a single fixed patch length for pretraining models on heterogeneous datasets~\cite{MOIRAI:woo2024unified, lagllama:rasul2023lag}.

The random approach independently samples a patch length $\ell$ from a configurable candidate pool $\mathcal{P}$ of integer patch lengths for each input. During training, we track the cumulative forecasting loss and usage count of each patch length over a validation epoch. Patch lengths with consistently poor performance—quantified via z-scores exceeding a threshold—are pruned from the pool. The pool is refined once per validation epoch, and the remaining lengths are reweighted inversely proportional to their squared mean validation loss, then normalized over the surviving lengths. This process gradually concentrates computation on lengths that empirically yield better predictions. Once the pool has been refined and the importance weights $w_{\ell}$ are normalized, the model obtains a forecast in one of two ways: (1) by selecting the best-performing length $\ell^* = \operatorname*{arg\,max}_{\ell} \; w_\ell$ and using its prediction, or (2) by computing a weighted ensemble forecast $\hat{x} = \sum_{\ell \in \mathcal{P}} w_{\ell} \hat{x}_{\ell}$. This approach is intended to minimize manual hyperparameter tuning and leverage patches of varying sizes to capture composite temporal patterns. All patch lengths share a single IVP Patcher with shared parameters.

\subsection{Time Normalization \& Domain Adaptation}\label{sec:timenorm_domainadapt}

Beyond normalizing values, we introduce a time normalization scheme to mitigate the impact of diverse temporal granularities. Models for regularly sampled series typically take only the value sequence as input and ignore timestamps, since adjacent samples share a fixed interval. With irregular data, however, gaps between observations vary and timestamps become essential. To prevent diverse time scales from interfering with the model's ability to learn temporal patterns, we first normalize timestamps before feeding them to the model.

Given a sequence of observed timestamps \(T = \{t_i\}_{i=1}^M\) associated with a sample from the training dataset, we define the dataset-level time unit \(\omega_g\) as the reciprocal of sampling frequency in the dataset, reflecting the recording resolution of the protocol. Intervals between consecutive observations are positive multiples of \(\omega_g\). For each sequence, we compute the time differences on the observed part, \(\{\Delta t_i = t_{i+1} - t_i\}_{i=1}^{M-1}\), and define the minimum sampling interval as the series-level unit \(\omega_s = \min_{1 \le i \le M-1} \Delta t_i\). Each sampling interval in the sequence is then scaled by dividing \(\omega_s\). The normalized timestamps are computed cumulatively \(t_i = \sum_{j=1}^{i-1} \Delta t_j / \omega_s\). After normalization, the smallest sampling interval among the observed timestamps is set to \(1\) (Fig.~\ref{fig:FlexTSF}(b)), and other sampling intervals are multiples of it. This produces unitless timestamps that unify different temporal granularities and irregularities. For forecast horizons, we normalize the future timestamps \(\{t_{M+h}\}_{h=1}^{H}\) using the same series-level unit \(\omega_s\) computed from the observed part. To summarize,
\begin{equation}
	\mathrm{TimeNorm}(t_i) := \frac{t_i - t_1}{\min_{1 \le j \le M-1} (t_{j+1}-t_j)}.
	\label{eq:timenorm}
\end{equation}
For supervised experiments with train-validation-test splits, we compute \(\omega_s\) on the training partition and apply Eq.~\eqref{eq:timenorm} to normalize timestamps in all partitions. The quantities \(\omega_g\) and \(\omega_s\) are also included in the static feature vector \(L\) for the Leader node in the next step.

To enable the model to self-adapt to diverse domains, we use the combined value- and time-normalization strategy to decouple static domain information from dynamic temporal patterns. Instead of leaving these statistics aside and using them to "de-normalize" output values of the forecasting model~\cite{wang2024timexer,luo2024deformabletst}, we concatenate the statistics into a feature vector $L = [\mu_{g}, \sigma_{g}, \mu_{s}, \sigma_{s}, \omega_{g}, \omega_{s}]$, where $\mu_{g}$ and $\sigma_{g}$ denote the dataset-level mean and standard deviation of the time series values, and $\mu_{s}$ and $\sigma_{s}$ denote the series-level mean and standard deviation of a time series. We add an extra computation node at the forefront of the causal self-attention (Fig.~\ref{fig:FlexTSF}(a)) that carries $L$ (referred to as the Leader node), which disseminates information to all subsequent nodes based on the attention's look-backward architecture. These six features are chosen as they are ubiquitous and can be obtained from any dataset. To integrate information, we append $K_{\text{out}}$ dummy patches to the end of the patch sequence as forecast placeholders, initialized with the input mean.

To enhance the self-attention mechanism's capability in capturing ordering and temporal information, we apply layer-wise time embedding, eliminating the need for input position embeddings. As self-attention has been found to be less effective in capturing the temporal order of time series~\cite{DLinear:zeng2023transformers, kim2024are}, we incorporate rotary position embedding (RoPE)~\cite{rope:su2024roformer} to embed each patch's start time $t_{k,0}$. Note that $\{t_{k,0}\}_{k=1}^{K}$ may exhibit variable intervals as the raw input time series can be irregular.

In FlexTSF, the start time $t_{k,0}$ (abbreviated as $t_k$ in the attention stacks and $t_0$ inside IVP Patcher) of each patch plays a crucial role in maintaining temporal coherence. Each patch is linked to its start time for time embedding in the attention stack. Within the IVP Patcher, backward solves align latent states to $t_{k,0}$, while forward solves propagate dynamics starting from it.

\section{Experiments}\label{sec:experiments}
\subsection{Experimental Setups}
We evaluate \approach{} across different domains and training setups using two non-overlapping groups of datasets: pre-training datasets $\mathcal{D}_{p}$ and held-out datasets $\mathcal{D}_{h}$. We conduct three stages of experiments: (i) time series forecasting using classic supervised train-validation-test splits on $\mathcal{D}_{h}$; (ii) zero-shot forecasting on the test partition of $\mathcal{D}_{h}$ with a model pre-trained on $\mathcal{D}_{p}$; and (iii) low-resource fine-tuning on $\mathcal{D}_{h}$ using a limited number of samples (e.g., 50, 100, 150).

\subsubsection{Datasets}

\begin{table*}[htbp]
	\footnotesize
	\centering
	\caption{Summary of Irregular Datasets in $\mathcal{D}_{h}$}
	\label{tab:irregular_dataset_summary}
	\setlength{\tabcolsep}{6pt}
	\begin{threeparttable}
		\begin{tabular}{l c r r ccc ccc ccc}
			\toprule
			\multirow{2}{*}{\textbf{Dataset}} & \multirow{2}{*}{\textbf{Irreg.}} & \multirow{2}{*}{\textbf{\#Var}} & \multirow{2}{*}{\textbf{$\omega_{g}$(s)}} & \multicolumn{3}{c}{\textbf{Missing rate (\%)}} & \multicolumn{3}{c}{\textbf{Input length}} & \multicolumn{3}{c}{\textbf{Sampling interval}}                                                                                             \\
			\cmidrule(lr){5-7}\cmidrule(lr){8-10}\cmidrule(lr){11-13}
			                                  &                                  &                                 &                                           & \textbf{Min}                                   & \textbf{Max}                              & \textbf{Mean}                                  & \textbf{Min} & \textbf{Max} & \textbf{Mean} & \textbf{Min} & \textbf{Max} & \textbf{Mean} \\
			\midrule
			SaTSM                             & $(2) (3)$                        & 13                              & 86400                                     & 0.00                                           & 0.00                                      & 0.00                                           & 8            & 48           & 26            & 1.00         & 335.00       & 12.91         \\
			MetrLA                            & $(1) (2)$                        & 207                             & 86400                                     & 0.00                                           & 63.65                                     & 3.95                                           & 4            & 12           & 11            & 3.5e-3       & 3.5e-3       & 3.5e-3        \\
			ArabDigit                         & $(1) (2)$                        & 13                              & 9.1e-5                                    & 5.39                                           & 14.25                                     & 10.00                                          & 4            & 41           & 19            & 1.00         & 1.00         & 1.00          \\
			CharTraj                          & $(1) (2)$                        & 3                               & 0.005                                     & 5.25                                           & 17.31                                     & 10.09                                          & 30           & 91           & 59            & 1.00         & 1.00         & 1.00          \\
			HAR                               & $(1) (3)$                        & 12                              & 0.1                                       & 74.00                                          & 75.00                                     & 75.00                                          & 25           & 25           & 25            & 1.00         & 17.00        & 3.31          \\
			eICU                              & $(1) (2) (3)$                    & 14                              & 60                                        & 29.80                                          & 85.90                                     & 65.13                                          & 30           & 146          & 57            & 1.00         & 851.00       & 24.83         \\
			Physio12                          & $(1) (2) (3)$                    & 37                              & 60                                        & 74.08                                          & 92.17                                     & 84.20                                          & 4            & 85           & 37            & 1.00         & 1220.00      & 38.15         \\
			MIMIC4                            & $(1) (2) (3)$                    & 71                              & 60                                        & 96.11                                          & 98.50                                     & 97.90                                          & 50           & 330          & 110           & 1.00         & 856.00       & 12.73         \\
			\bottomrule
		\end{tabular}
		\begin{tablenotes}[para]
			\scriptsize
			\noindent Irreg.\ indicates irregularity types: (1) missing values, (2) varying sequence lengths, and (3) unequal sampling intervals. $\omega_{g}$ is the dataset-level time unit (seconds) and sampling intervals are reported as multiples of $\omega_{g}$.
		\end{tablenotes}
	\end{threeparttable}
\end{table*}

Our pre-training group $\mathcal{D}_{p}$ consists of 2.4 million sequences with lengths ranging from 18 to 1024, totaling 1.2 billion observations. The overall missing rate is 3.63\%, and no imputation is applied. The data are sourced from the Monash~\cite{Monash:godahewa2021monash} and UCR\&UEA~\cite{UCR:dau2019ucr, UEA:bagnall2018uea, timeseriesclassification} repositories, covering diverse domains such as climate, health, banking, tourism, etc. The held-out group $\mathcal{D}_{h}$ contains datasets from the Long Time Series Forecasting Benchmark~\cite{LTF:lai2018modeling, Autoformer:Wu2021AutoformerDT} and the Irregular Time Series Benchmark~\cite{latentode:rubanova2019latent, DCRNN:li2018diffusion}. After removing overlaps with $\mathcal{D}_{p}$, $\mathcal{D}_{h}$ contains $8$ commonly used datasets with regular time series (ETTh1, ETTh2, ETTm1, ETTm2, Illness, ExRate, Weather, Electricity)~\cite{Autoformer:Wu2021AutoformerDT} and $8$ datasets with time series of different irregularities. Each dataset in $\mathcal{D}_{h}$ is split into training, validation, and testing sets, following their original splits if known or a split ratio of 8:1:1, otherwise.

For the irregular group, Table~\ref{tab:irregular_dataset_summary} lists the 8 datasets along with types of temporal irregularity they exhibit. The missing rate, input length, and sampling interval are computed on the test partition of each dataset as this partition is used consistently by classic, zero-shot and fine-tuning experiments. SaTSM is a Sentinel-2 Earth observation time series with 13 spectral bands for land-cover monitoring\footnote{\url{https://zenodo.org/records/5712933}}. MetrLA contains traffic speed readings from loop sensors across the Los Angeles freeway network\footnote{\url{https://github.com/liyaguang/DCRNN}}. ArabDigit and CharTraj datasets are from UEA time series archive~\cite{UEA:bagnall2018uea}. They mainly exhibit variable sequence lengths with limited missingness. We randomly dropped 10\% of the observations in these datasets to increase the difficulty of model evaluation. HAR (Human Activity Recognition) provides multivariate wearable sensor streams recording 3D motion data from individuals~\cite{latentode:rubanova2019latent}. eICU\footnote{\url{https://physionet.org/content/eicu-crd/2.0/}}, Physio12 (PhysioNet 2012 Challenge)\footnote{\url{https://physionet.org/content/challenge-2012/1.0.0/}}, and MIMIC4\footnote{\url{https://physionet.org/content/mimiciv/1.0/}} are electronic health record (EHR) datasets comprising multivariate clinical time series of vital signs, laboratory results, and interventions, and are particularly challenging due to their high missing rates, wide variation in sequence length, and irregular sampling intervals.

\subsubsection{Baselines}

We incorporate a comprehensive set of baselines spanning three categories: (1) \textbf{Regular}: models assuming uniformly spaced and complete time series, including DLinear~\cite{DLinear:zeng2023transformers}, NHiTS~\cite{NHITS:challu2023nhits}, PatchTST~\cite{PatchTST:nie2023time}, iTransformer~\cite{itransformer:liu2023itransformer}, and TimeMixer~\cite{TimeMixer:wang2024timemixer}; (2) \textbf{Irregular}: models explicitly handling temporal irregularities, including mTAN~\cite{mTAN:Shukla2021mTAN}, Latent-Flow~\cite{Bilos2021:neural-flow}, CRU~\cite{cru:schirmer2022modeling}, ContiFormer~\cite{Contiformer:chen2023}, IVP-VAE~\cite{ivpvae:xiao2024ivp}, and t-PatchGNN~\cite{t-PatchGNN:zhang2024irregular}; and (3) \textbf{Zero-shot}: universal forecasters pre-trained on large corpora, including Lag-Llama~\cite{lagllama:rasul2023lag}, TimesFM~\cite{TimesFM:das2024decoder}, MOIRAI-MoE~\cite{Moirai-MoE:liu2025}, Sundial~\cite{Sundial:liu2025}, and the statistical model ARIMA~\cite{arima:box1970time} as a reference. Since baselines designed for regular time series and zero-shot forecasting do not natively support irregular data, we adapt them by imputing missing values using the mean of each time series. For all baselines, we adopt the publicly released architectures and default hyperparameters from the original implementations and, when necessary, adjust optimization settings (e.g., learning rate and batch size) via grid search within the recommended ranges to adapt them to new datasets.

\begin{table*}[!t]
	\centering
	\caption{Forecasting Performance in MSE on Irregular Time Series Datasets With Different Forecast Lengths}
	\label{tab:tvt_irregular_results}
	\begingroup
	\footnotesize
	\setlength{\tabcolsep}{5pt}
	\renewcommand{\arraystretch}{1.2}
	\begin{threeparttable}
		\begin{tabular}{llcccccccc}
			\toprule
			                              & $\nicefrac{H}{M}$ & SaTSM                       & MetrLA                      & ArabDigit                   & CharTraj                    & HAR                         & eICU                        & Physio12                    & MIMIC4                      \\
			\midrule
			\multicolumn{10}{l}{\emph{Regular}}                                                                                                                                                                                                                                                               \\
			\midrule
			\multirow{3}{*}{DLinear}      & 0.25              & 0.294$\pm$0.001             & 0.365$\pm$0.000             & 0.825$\pm$0.000             & 0.479$\pm$0.001             & 0.621$\pm$0.066             & 0.658$\pm$0.008             & 0.563$\pm$0.042             & 0.733$\pm$0.002             \\
			                              & 0.50              & 0.372$\pm$0.001             & 0.448$\pm$0.000             & 0.928$\pm$0.001             & 0.615$\pm$0.001             & 0.678$\pm$0.018             & 0.671$\pm$0.005             & 0.599$\pm$0.042             & 0.752$\pm$0.017             \\
			                              & 1.00              & 0.424$\pm$0.000             & 0.571$\pm$0.000             & 0.905$\pm$0.001             & 0.773$\pm$0.003             & 0.686$\pm$0.027             & 0.714$\pm$0.007             & 0.622$\pm$0.028             & 0.802$\pm$0.015             \\
			\midrule
			\multirow{3}{*}{NHiTS}        & 0.25              & 0.488$\pm$0.002             & 1.051$\pm$0.002             & 0.925$\pm$0.001             & 0.262$\pm$0.009             & 0.934$\pm$0.009             & 0.976$\pm$0.029             & 0.855$\pm$0.053             & 0.906$\pm$0.009             \\
			                              & 0.50              & 0.498$\pm$0.002             & 1.041$\pm$0.001             & 0.963$\pm$0.002             & 0.399$\pm$0.000             & 0.930$\pm$0.003             & 0.951$\pm$0.020             & 0.867$\pm$0.052             & 0.903$\pm$0.009             \\
			                              & 1.00              & 0.546$\pm$0.003             & 1.049$\pm$0.001             & 0.912$\pm$0.003             & 0.403$\pm$0.006             & 0.813$\pm$0.006             & 0.938$\pm$0.011             & 0.862$\pm$0.038             & 0.924$\pm$0.004             \\
			\midrule
			\multirow{3}{*}{PatchTST}     & 0.25              & 0.277$\pm$0.007             & 0.356$\pm$0.001             & 0.730$\pm$0.002             & 0.318$\pm$0.008             & 0.252$\pm$0.020             & 0.581$\pm$0.012             & 0.546$\pm$0.043             & 0.661$\pm$0.015             \\
			                              & 0.50              & 0.354$\pm$0.001             & 0.450$\pm$0.000             & 0.865$\pm$0.003             & 0.440$\pm$0.004             & 0.285$\pm$0.017             & 0.610$\pm$0.003             & 0.587$\pm$0.046             & 0.693$\pm$0.007             \\
			                              & 1.00              & 0.363$\pm$0.001             & 0.601$\pm$0.000             & 0.872$\pm$0.003             & 0.454$\pm$0.009             & 0.311$\pm$0.004             & 0.669$\pm$0.005             & 0.607$\pm$0.026             & 0.752$\pm$0.010             \\
			\midrule
			\multirow{3}{*}{iTransformer} & 0.25              & 0.270$\pm$0.003             & \underline{0.345$\pm$0.001} & 0.617$\pm$0.002             & 0.172$\pm$0.009             & 0.227$\pm$0.026             & 0.564$\pm$0.013             & 0.520$\pm$0.034             & 0.686$\pm$0.005             \\
			                              & 0.50              & 0.341$\pm$0.001             & \underline{0.428$\pm$0.000} & 0.732$\pm$0.003             & 0.257$\pm$0.010             & 0.241$\pm$0.030             & 0.595$\pm$0.009             & 0.557$\pm$0.037             & 0.709$\pm$0.010             \\
			                              & 1.00              & \textbf{0.353$\pm$0.001}    & 0.542$\pm$0.001             & 0.753$\pm$0.003             & 0.319$\pm$0.009             & \underline{0.256$\pm$0.009} & 0.658$\pm$0.005             & 0.589$\pm$0.018             & 0.764$\pm$0.010             \\
			\midrule
			\multirow{3}{*}{TimeMixer}    & 0.25              & 0.278$\pm$0.003             & 0.352$\pm$0.000             & 0.732$\pm$0.016             & 0.372$\pm$0.008             & 0.478$\pm$0.035             & 0.557$\pm$0.017             & 0.489$\pm$0.040             & 0.635$\pm$0.006             \\
			                              & 0.50              & 0.355$\pm$0.000             & 0.443$\pm$0.001             & 0.856$\pm$0.009             & 0.483$\pm$0.008             & 0.568$\pm$0.026             & 0.590$\pm$0.014             & 0.527$\pm$0.042             & 0.669$\pm$0.018             \\
			                              & 1.00              & 0.375$\pm$0.001             & 0.590$\pm$0.001             & 0.854$\pm$0.003             & 0.552$\pm$0.015             & 0.630$\pm$0.020             & 0.647$\pm$0.006             & 0.561$\pm$0.031             & 0.725$\pm$0.016             \\
			\midrule
			\multicolumn{10}{l}{\emph{Irregular}}                                                                                                                                                                                                                                                             \\
			\midrule
			\multirow{3}{*}{mTAN}         & 0.25              & 0.345$\pm$0.053             & 0.400$\pm$0.002             & 0.568$\pm$0.028             & 0.347$\pm$0.004             & 0.298$\pm$0.016             & \underline{0.482$\pm$0.015} & \underline{0.457$\pm$0.047} & \underline{0.593$\pm$0.017} \\
			                              & 0.50              & 0.385$\pm$0.021             & 0.435$\pm$0.002             & 0.677$\pm$0.002             & 0.476$\pm$0.010             & 0.307$\pm$0.011             & \underline{0.530$\pm$0.011} & \underline{0.500$\pm$0.043} & \underline{0.635$\pm$0.003} \\
			                              & 1.00              & 0.448$\pm$0.004             & 0.498$\pm$0.007             & 0.720$\pm$0.018             & 0.586$\pm$0.015             & 0.323$\pm$0.015             & \underline{0.600$\pm$0.008} & \underline{0.532$\pm$0.030} & \underline{0.706$\pm$0.014} \\
			\midrule
			\multirow{3}{*}{Latent-Flow}  & 0.25              & 0.568$\pm$0.038             & 0.403$\pm$0.014             & \textbf{0.543$\pm$0.002}    & 0.383$\pm$0.059             & 0.350$\pm$0.010             & 0.620$\pm$0.055             & 0.646$\pm$0.000             & 0.909$\pm$0.078             \\
			                              & 0.50              & 0.502$\pm$0.013             & 0.452$\pm$0.009             & 0.676$\pm$0.005             & \underline{0.250$\pm$0.012} & 0.342$\pm$0.001             & 0.582$\pm$0.028             & 0.772$\pm$0.104             & 0.965$\pm$0.094             \\
			                              & 1.00              & 0.542$\pm$0.038             & 0.630$\pm$0.027             & \underline{0.712$\pm$0.003} & \underline{0.300$\pm$0.001} & 0.343$\pm$0.003             & 0.626$\pm$0.052             & 0.810$\pm$0.109             & 0.973$\pm$0.141             \\
			\midrule
			\multirow{3}{*}{CRU}          & 0.25              & \textbf{0.262$\pm$0.003}    & 0.392$\pm$0.000             & 0.553$\pm$0.004             & 0.210$\pm$0.007             & 0.463$\pm$0.009             & 0.557$\pm$0.016             & 0.530$\pm$0.049             & 0.672$\pm$0.012             \\
			                              & 0.50              & \underline{0.338$\pm$0.002} & 0.433$\pm$0.001             & 0.678$\pm$0.007             & 0.463$\pm$0.017             & 0.517$\pm$0.014             & 0.621$\pm$0.025             & 0.581$\pm$0.049             & 0.720$\pm$0.005             \\
			                              & 1.00              & 0.375$\pm$0.006             & \underline{0.491$\pm$0.003} & 0.725$\pm$0.010             & 0.636$\pm$0.009             & 0.643$\pm$0.011             & 0.726$\pm$0.015             & 0.608$\pm$0.025             & 0.783$\pm$0.019             \\
			\midrule
			\multirow{3}{*}{ContiFormer}  & 0.25              & 0.440$\pm$0.055             & 0.384$\pm$0.001             & 0.546$\pm$0.010             & 0.298$\pm$0.034             & 0.443$\pm$0.021             & 0.859$\pm$0.049             & 0.758$\pm$0.025             & 0.933$\pm$0.008             \\
			                              & 0.50              & 0.449$\pm$0.013             & 0.436$\pm$0.002             & 0.683$\pm$0.015             & 0.516$\pm$0.044             & 0.450$\pm$0.010             & 0.865$\pm$0.025             & 0.762$\pm$0.029             & 0.915$\pm$0.001             \\
			                              & 1.00              & 0.547$\pm$0.028             & 0.612$\pm$0.005             & 0.741$\pm$0.009             & 0.644$\pm$0.024             & 0.453$\pm$0.013             & 0.868$\pm$0.005             & 0.764$\pm$0.019             & 0.920$\pm$0.000             \\
			\midrule
			\multirow{3}{*}{IVP-VAE}      & 0.25              & 0.394$\pm$0.035             & 0.415$\pm$0.002             & 0.550$\pm$0.003             & \underline{0.160$\pm$0.002} & 0.222$\pm$0.012             & 0.716$\pm$0.020             & 0.543$\pm$0.043             & 0.865$\pm$0.040             \\
			                              & 0.50              & 0.423$\pm$0.004             & 0.449$\pm$0.002             & \underline{0.674$\pm$0.004} & 0.305$\pm$0.006             & 0.247$\pm$0.012             & 0.615$\pm$0.016             & 0.593$\pm$0.045             & 0.899$\pm$0.068             \\
			                              & 1.00              & 0.519$\pm$0.004             & 0.498$\pm$0.005             & 0.728$\pm$0.004             & 0.394$\pm$0.004             & 0.275$\pm$0.011             & 0.752$\pm$0.090             & 0.774$\pm$0.018             & 0.967$\pm$0.080             \\
			\midrule
			\multirow{3}{*}{t-PatchGNN}   & 0.25              & 0.347$\pm$0.018             & 0.422$\pm$0.009             & 0.759$\pm$0.017             & 0.553$\pm$0.013             & \underline{0.202$\pm$0.023} & 0.781$\pm$0.016             & 0.632$\pm$0.042             & 0.901$\pm$0.065             \\
			                              & 0.50              & 0.411$\pm$0.005             & 0.472$\pm$0.012             & 0.825$\pm$0.025             & 0.716$\pm$0.026             & \underline{0.239$\pm$0.018} & 0.819$\pm$0.051             & 0.705$\pm$0.048             & 0.925$\pm$0.070             \\
			                              & 1.00              & 0.481$\pm$0.004             & 0.513$\pm$0.019             & 0.974$\pm$0.018             & 0.809$\pm$0.020             & 0.259$\pm$0.017             & 0.875$\pm$0.044             & 0.765$\pm$0.015             & 0.967$\pm$0.085             \\
			\midrule
			\multicolumn{10}{l}{\emph{Ours}}                                                                                                                                                                                                                                                                  \\
			\midrule
			\multirow{3}{*}{FlexTSF}      & 0.25              & \textbf{0.262$\pm$0.003}    & \textbf{0.330$\pm$0.001}    & \underline{0.545$\pm$0.002} & \textbf{0.090$\pm$0.003}    & \textbf{0.192$\pm$0.024}    & \textbf{0.467$\pm$0.014}    & \textbf{0.384$\pm$0.014}    & \textbf{0.479$\pm$0.012}    \\
			                              & 0.50              & \textbf{0.334$\pm$0.001}    & \textbf{0.426$\pm$0.001}    & \textbf{0.673$\pm$0.004}    & \textbf{0.191$\pm$0.008}    & \textbf{0.218$\pm$0.012}    & \textbf{0.512$\pm$0.009}    & \textbf{0.435$\pm$0.022}    & \textbf{0.549$\pm$0.007}    \\
			                              & 1.00              & \underline{0.355$\pm$0.002} & \textbf{0.488$\pm$0.005}    & \textbf{0.711$\pm$0.009}    & \textbf{0.254$\pm$0.006}    & \textbf{0.252$\pm$0.013}    & \textbf{0.589$\pm$0.007}    & \textbf{0.493$\pm$0.016}    & \textbf{0.657$\pm$0.012}    \\
			\bottomrule
		\end{tabular}

		\begin{tablenotes}[para]
			\scriptsize
			\noindent $\nicefrac{H}{M}$ indicates the ratio of forecasting horizon ($H$) to input length ($M$). The best result across methods is in \textbf{bold}, and the second best is underlined.
		\end{tablenotes}
	\end{threeparttable}
	\endgroup
\end{table*}

\subsubsection{Implementation Details}

The pre-trained \approach{} model consists of 6 attention layers, each with 12 self-attention heads and a hidden size of 768, for a total of 63 million parameters. In the IVP Patcher, we employ ResNetFlow~\cite{Bilos2021:neural-flow} to solve initial value problems. Regarding patch length, we used the random approach (see Section~\ref{subsec:random_patches}) in classic supervised experiments and the deterministic approach in pre-training, zero-shot and fine-tuning. For the random approach, we initialize $\mathcal{P} = \{1, 2, \ldots, 16\}$, set the z-score pruning threshold to $1.3$, and use the weighted-ensemble rule $\hat{x} = \sum_{\ell} w_{\ell}\,\hat{x}_{\ell}$ at inference. The Leader node of FlexTSF is implemented as a multilayer perceptron. We set the loss function coefficients to $\lambda = 0.1$ and $\beta = 1.0$. We use Adam with a stepwise learning-rate decay (initial learning rate $10^{-4}$, decay factor $0.5$ every 20 epochs) for classic training and low-resource fine-tuning, and AdamW with cosine-annealing and 1{,}000 warmup steps for pre-training, all with batch size 64 and moderate weight decay. We found that varying batch size between 32 and 128 or scaling the learning rate within a small range yields similar performance and does not affect the conclusions of our empirical study.

For regular datasets, we use a fixed input length of 96 and a forecasting horizon of 96. For irregular datasets, it is impractical to define fixed input and output lengths across all cases. Instead, we adopt a forecast-to-input ratio $\nicefrac{H}{M}$ and vary it across $\{0.25, 0.5, 1.0\}$ to evaluate performance under different forecasting horizons.
Following prior work~\cite{Chronos:ansari2024chronos,PatchTST:nie2023time,mTAN:Shukla2021mTAN}, each experiment is repeated three times with different random seeds for dataset splitting and parameter initialization. Model performance is reported as the mean and standard deviation of Mean Squared Error (MSE). We choose MSE because it matches our L2 training objective and yields stable, comparable evaluations, whereas RMSE is a monotonic re-parameterization and percentage errors (e.g., MAPE) are unstable near small or zero targets. All models are evaluated under the same computing environment with NVIDIA A100 GPUs for classic training, and H100 GPUs for pre-training and zero-shot inference.

\subsection{Time Series Forecasting}\label{sec:exp_forecasting}

Table~\ref{tab:tvt_irregular_results} presents the results in the classic training-validation-testing setting on held-out irregular time series datasets. \approach{} consistently outperforms other models in $22$ of $24$ cases (8 datasets × 3 horizons), demonstrating its effectiveness in capturing temporal dynamics and dealing with irregularities. Among the baselines, some methods originally designed for regular time series achieve decent results on irregular datasets, while those tailored for irregular time series perform particularly well on datasets characterized by higher rates of missing data and greater sparsity, such as eICU, Physio12 and MIMIC4.

\begin{figure}[htbp]
	\centering
	\includegraphics[width=1.0\linewidth]{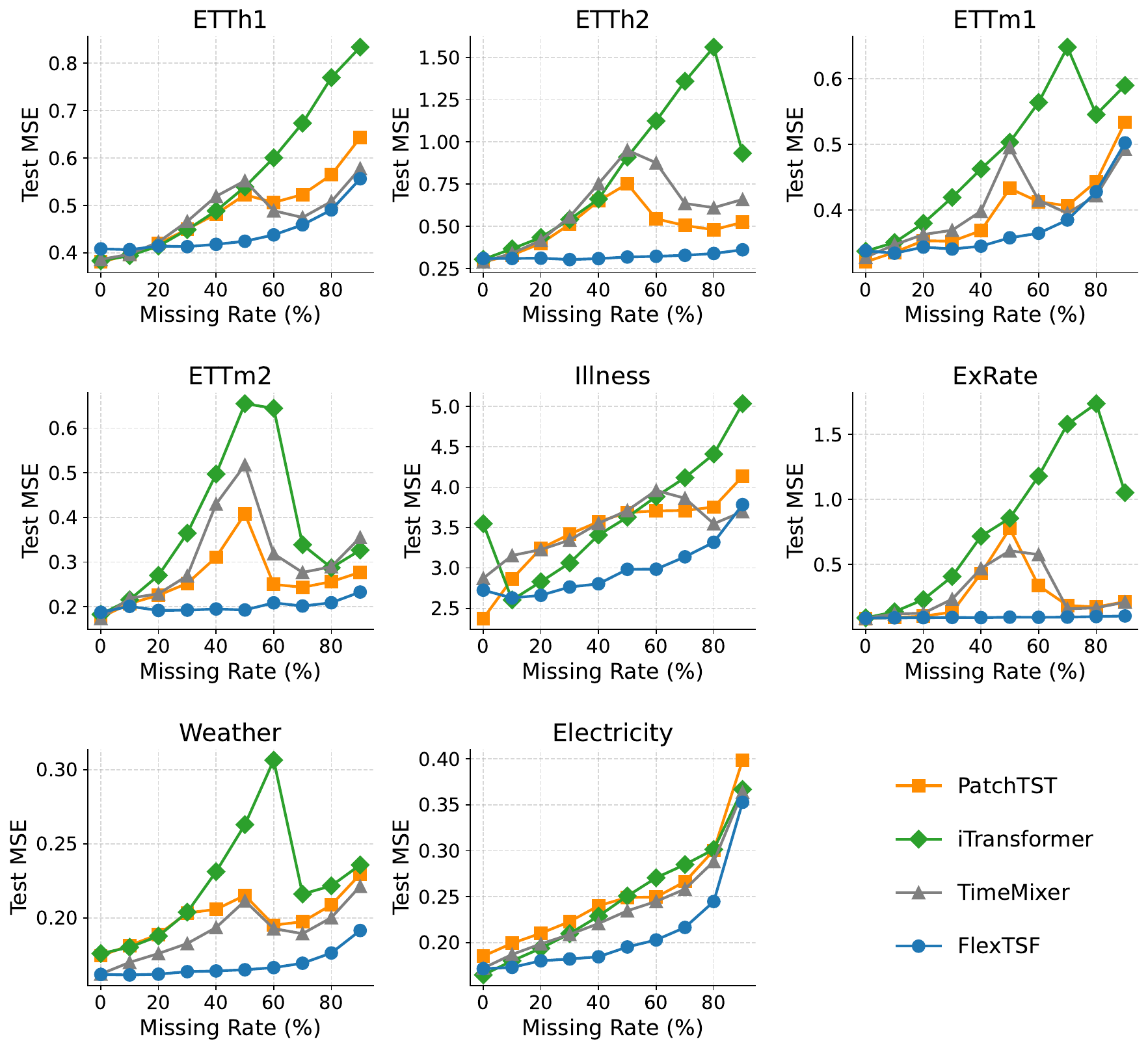}
	\caption{Test MSE of four representative models under varying missing rates. A missing rate of 0\% indicates fully observed regular time series.}
	\label{fig:missing_rate_comparison}
\end{figure}

Fig.~\ref{fig:missing_rate_comparison} further investigates the impact of various missing rates on the forecasting performance of four representative models. As expected, model performance degrades (higher MSE) as the missing rate increases. However, \approach{} consistently exhibits superior robustness, maintaining a relatively lower MSE across missing rates and datasets, highlighting its effectiveness in handling irregular data. The MSE of models like PatchTST and TimeMixer initially rises and then decreases at very high missing rates. This behavior occurs because, at high missing rates, models may shift focus to the remaining patterns, and heavy imputation can smooth out noisy fluctuations they might otherwise overfit to.

\subsection{Zero-shot and Low-resource Fine-tuning}
Table~\ref{table:zeroshot} shows the results of our zero-shot experiments, where we pre-train \approach{} on $\mathcal{D}_{p}$ and then directly apply it to the test sets of held-out irregular time series datasets. \approach{} achieves the best performance on six out of the eight benchmarks and the second-best on the remaining two, MetrLA and ArabDigit, highlighting its potential as a universal forecasting model.

\begin{table}[ht]
	\footnotesize
	\centering
	\caption{Zero-Shot Forecasting Performance Comparison Between \approach{} and Other Pre-Trained Models on Irregular Time Series Datasets}
	\begingroup
	\setlength{\tabcolsep}{4pt}
	\begin{tabular}{@{}lcccccc@{}}
		\toprule
		Dataset   & {\scriptsize Lag‐Llama} & {\scriptsize TimesFM} & {\scriptsize MOIRAI-MoE} & {\scriptsize Sundial} & {\scriptsize ARIMA} & {\scriptsize FlexTSF} \\
		\midrule
		SaTSM     & 0.946                   & 0.900                 & 0.913                    & \underline{0.684}     & 0.895               & \textbf{0.675}        \\
		MetrLA    & 0.691                   & 0.645                 & 0.563                    & \textbf{0.365}        & 0.771               & \underline{0.420}     \\
		ArabDigit & 0.947                   & 1.038                 & 1.076                    & \textbf{0.767}        & 0.943               & \underline{0.881}     \\
		CharTraj  & \underline{0.513}       & 0.908                 & 0.637                    & 0.541                 & 0.619               & \textbf{0.414}        \\
		HAR       & 0.464                   & 0.681                 & \underline{0.350}        & 0.550                 & 0.685               & \textbf{0.242}        \\
		eICU      & 0.710                   & 0.915                 & 0.878                    & \underline{0.604}     & 0.944               & \textbf{0.593}        \\
		Physio12  & 0.747                   & 1.052                 & 0.795                    & \underline{0.515}     & 0.834               & \textbf{0.492}        \\
		MIMIC4    & 0.725                   & 0.930                 & 0.893                    & \underline{0.687}     & 0.918               & \textbf{0.608}        \\
		\bottomrule
	\end{tabular}
	\endgroup
	\label{table:zeroshot}
\end{table}

\begin{figure}[htbp]
	\centering
	\includegraphics[width=1.0\linewidth]{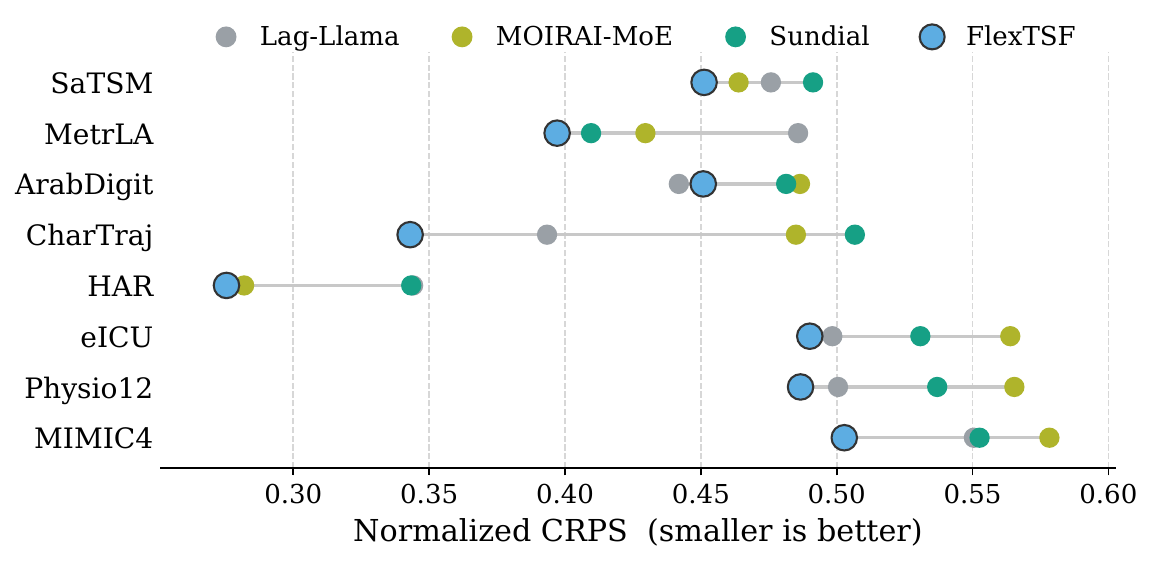}
	\caption{Zero-shot probabilistic forecasting quality, measured by
		normalized CRPS (smaller is better), on the eight irregular
		benchmarks under the same setting as Table~\ref{table:zeroshot}.}
	\label{fig:zeroshot_crps}
\end{figure}

To further assess probabilistic forecasting quality beyond point accuracy, we compare \approach{} against the three probabilistic foundation models in Table~\ref{table:zeroshot} (Lag-Llama, MOIRAI-MoE, and Sundial) using normalized CRPS, a metric jointly sensitive to calibration and sharpness, computed per channel, normalized by that channel's standard deviation, and averaged across channels. As shown in Fig.~\ref{fig:zeroshot_crps}, \approach{} attains the lowest normalized CRPS on seven of the eight benchmarks, with ArabDigit the only exception, where it is marginally behind Lag-Llama. The MetrLA and ArabDigit results admit a coherent reading across both Table~\ref{table:zeroshot} and Fig.~\ref{fig:zeroshot_crps}. Both datasets have constant sampling intervals and relatively low missing rates (Table~\ref{tab:irregular_dataset_summary}), a regime in which \approach{}'s continuous-time IVP Patcher offers less marginal benefit. However, the MSE picture does not extend to the calibration-sensitive metric: \approach{} reclaims the lead on MetrLA, and on ArabDigit Sundial drops below both Lag-Llama and \approach{}. Across Fig.~\ref{fig:zeroshot_crps}, Sundial's advantage over the other foundation models likewise diminishes under CRPS, consistent with strong point estimates not always translating to well-calibrated predictive distributions.

\begin{figure}[htbp]
	\centering
	\includegraphics[width=1.0\linewidth]{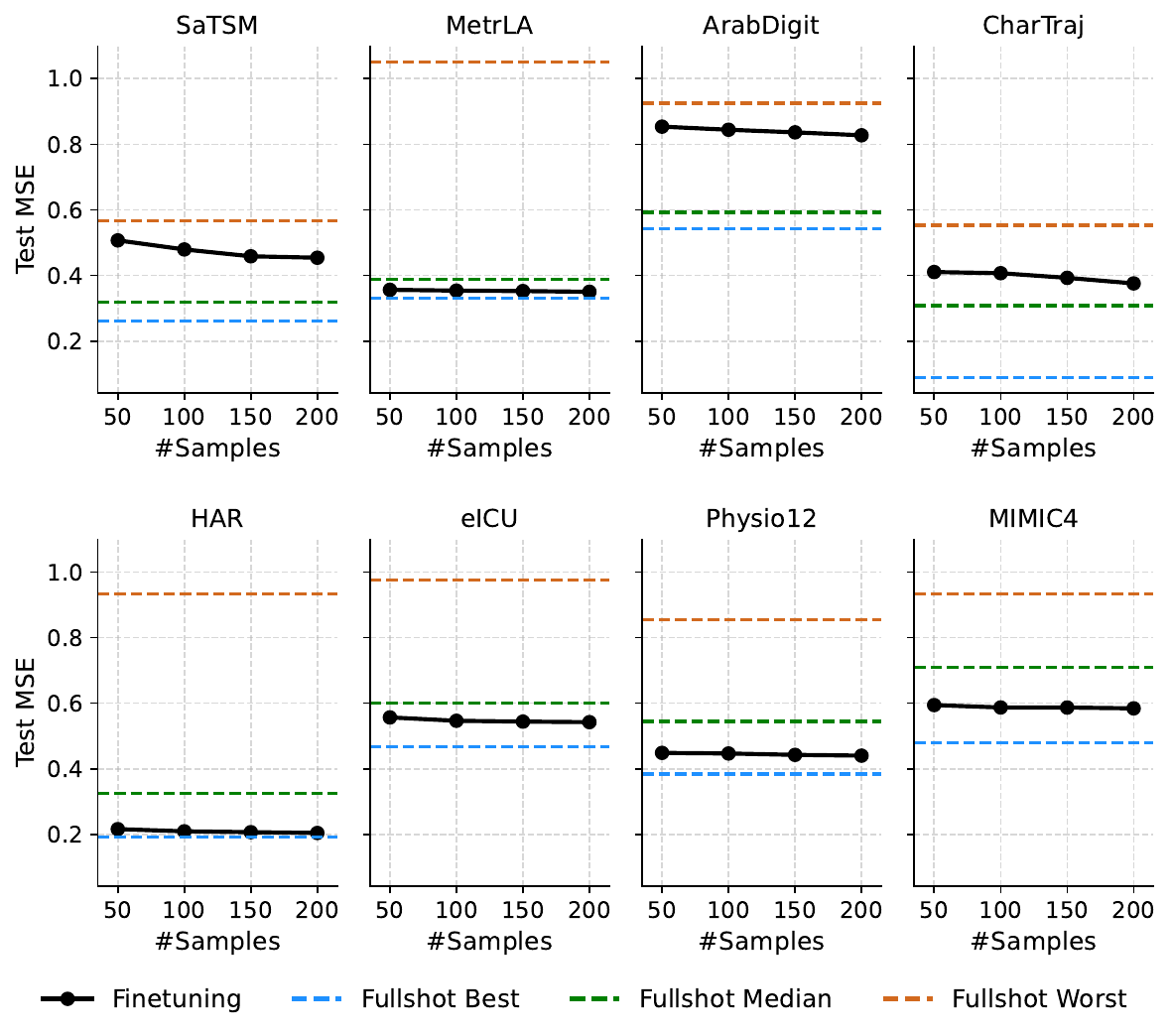}
	\caption{Results of the low-resource fine-tuning experiments. The x-axis shows the number of samples used for fine-tuning. The best, median, and worst full-shot training results from Section \ref{sec:exp_forecasting} are shown on the plots for reference.}
	\label{fig:fewshot_exps}
\end{figure}

After pre-training on $\mathcal{D}_{p}$, \approach{} demonstrates effective performance by fine-tuning only the Leader node (a small subset of parameters) on target domains while keeping the remaining model parameters frozen, using varying numbers of samples drawn from the training sets of $\mathcal{D}_{h}$. Fig.~\ref{fig:fewshot_exps} presents the results. For reference, the best, median, and worst outcomes from full-shot training (Table~\ref{tab:tvt_irregular_results}) are also shown. \approach{} consistently outperforms the worst full-shot results, exceeds the median on five datasets, and matches the best on two. This capability reduces the time and computational resources required for domain adaptation, making it effective in scenarios with limited training resources.

\subsection{Ablation Study} \label{sec:ablation}
To assess each component, we independently remove the IVP Patcher, time normalization, and the Leader node. Without the IVP Patcher, we use a linear projection after filling missing entries with the mean value of each sequence. Without time normalization, timestamps are used without series-level rescaling. Without the Leader node, both the static statistics features and the leader-node module are omitted. Table~\ref{table:abla_nonorm} reports the MSE changes relative to the full \approach{} model in zero-shot scenarios.

\begin{table}[htbp]
	\footnotesize
	\centering
	\caption{Ablation of \approach{} Components}
	\begingroup
	\setlength{\tabcolsep}{5pt}
	\begin{threeparttable}
		\begin{tabular}{@{}lcccccc@{}}
			\toprule
			          & \multicolumn{2}{c}{IVP Patcher} & \multicolumn{2}{c}{Time Norm} & \multicolumn{2}{c}{Leader Node}                               \\
			\cmidrule(lr){2-3}\cmidrule(lr){4-5}\cmidrule(lr){6-7}
			Dataset   & MSE                             & MSE+/-                        & MSE                             & MSE+/-   & MSE   & MSE+/-   \\
			\midrule
			SaTSM     & 0.703                           & +4.15\%                       & 0.676                           & +0.15\%  & 0.688 & +1.93\%  \\
			MetrLA    & 0.459                           & +9.29\%                       & 0.486                           & +15.71\% & 0.466 & +10.95\% \\
			ArabDigit & 0.908                           & +3.06\%                       & 0.888                           & +0.79\%  & 0.944 & +7.15\%  \\
			CharTraj  & 0.544                           & +31.40\%                      & 0.535                           & +29.23\% & 0.519 & +25.36\% \\
			HAR       & 0.291                           & +20.25\%                      & 0.245                           & +1.24\%  & 0.248 & +2.48\%  \\
			eICU      & 0.630                           & +6.24\%                       & 0.652                           & +9.95\%  & 0.614 & +3.54\%  \\
			Physio12  & 0.548                           & +11.38\%                      & 0.528                           & +7.32\%  & 0.500 & +1.63\%  \\
			MIMIC4    & 0.626                           & +2.96\%                       & 0.652                           & +7.24\%  & 0.638 & +4.93\%  \\
			Average   & 0.589                           & +11.09\%                      & 0.583                           & +8.95\%  & 0.577 & +7.25\%  \\
			\bottomrule
		\end{tabular}

		\begin{tablenotes}[para]
			\scriptsize
			\noindent MSE+/- indicates the change relative to \approach{}’s zero-shot results in Table~\ref{table:zeroshot}.
		\end{tablenotes}
	\end{threeparttable}

	\endgroup
	\label{table:abla_nonorm}
\end{table}

Removing the IVP Patcher increases average MSE by 11.09\%, showing the benefit of latent continuous-time patching over imputation-based linear patching. Removing time normalization increases MSE by 8.95\%, confirming its role in unifying heterogeneous temporal granularities for cross-domain transfer. Removing the Leader node increases MSE by 7.25\%. This indicates that the static statistics aid domain adaptation, while the smaller degradation suggests that \approach{} still primarily relies on the temporal modeling components.

\subsection{Impact of Random Patch Lengths}

\begin{figure}[htbp]
	\centering
	\includegraphics[width=0.9\linewidth]{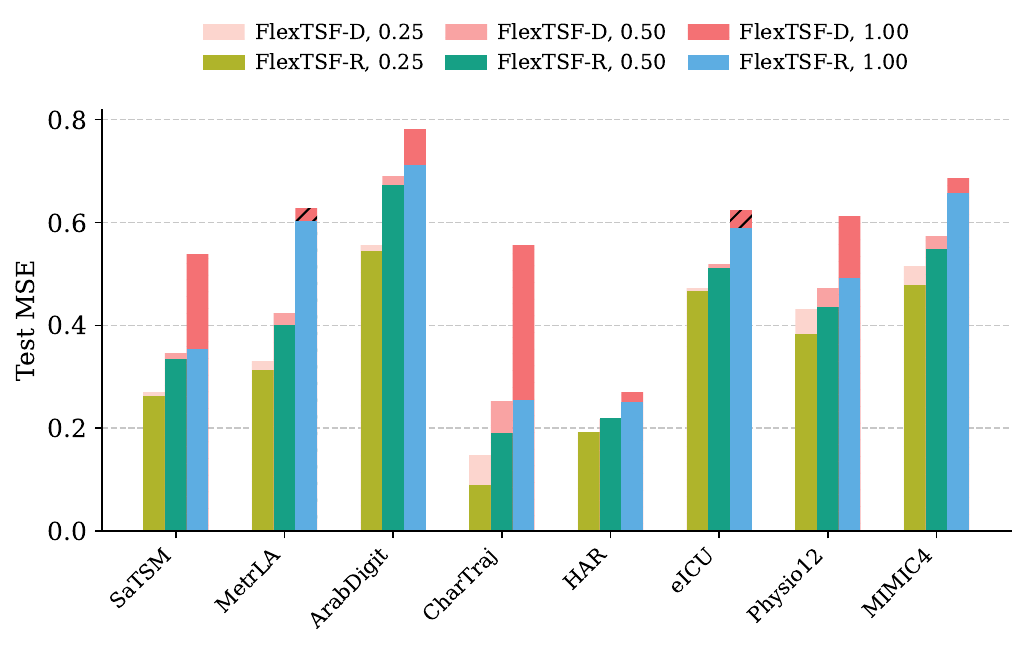}
	\caption{Comparison of models trained by random patch lengths (FlexTSF-R) and deterministic patch lengths (FlexTSF-D) across different $\nicefrac{H}{M}$ values ($0.25$, $0.50$, $1.00$).}
	\label{fig:flex_rplw_comparison}
\end{figure}

To investigate the impact of the random patch length strategy (Section~\ref{subsec:random_patches}), Fig.~\ref{fig:flex_rplw_comparison} compares models trained with random patch lengths and deterministic patch lengths. Across all datasets and $\nicefrac{H}{M}$ values, \approach{}-R consistently achieves lower test MSE than \approach{}-D. The performance gap becomes especially pronounced when the forecast length is larger (i.e., $\nicefrac{H}{M} = 1.00$), highlighting the advantage of randomization and multi-patch composition in capturing long-term temporal patterns.

\begin{figure}[htbp]
	\centering
	\includegraphics[width=0.95\linewidth]{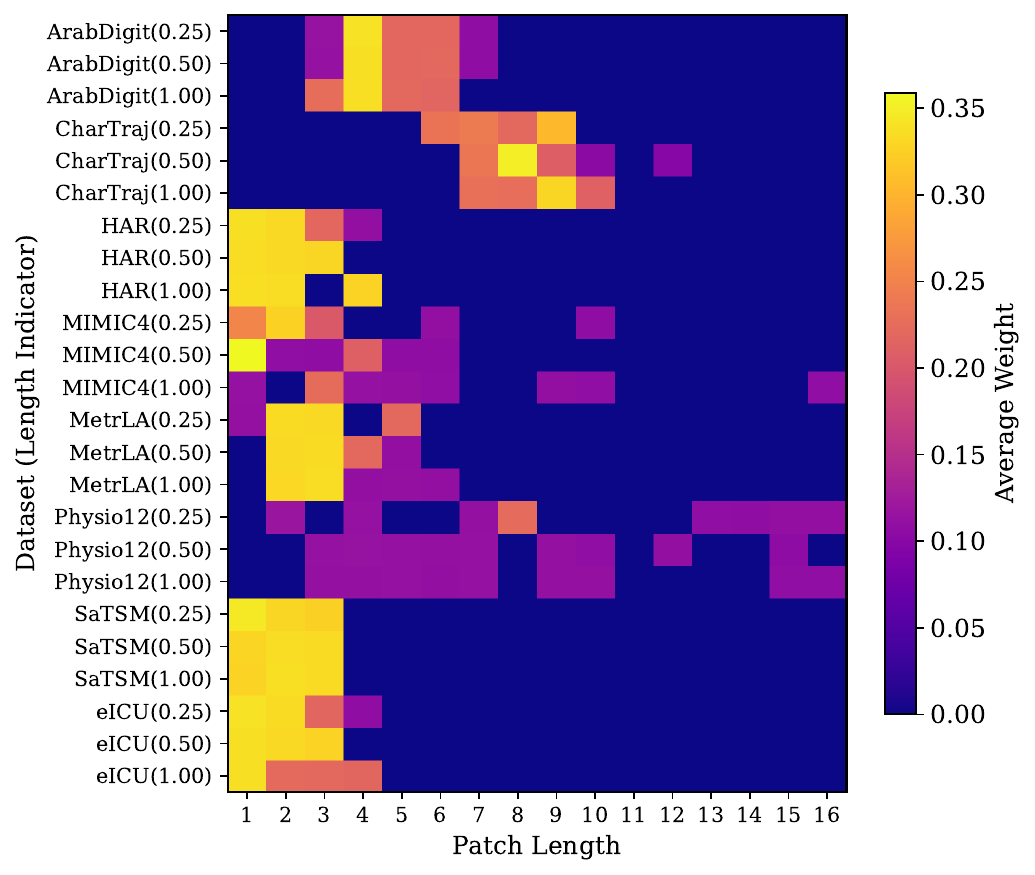}
	\caption{Patch length distribution after dynamic selection. Each row corresponds to a dataset and forecast-ratio setting.}
	\label{fig:heatmap_patchlen_irregular}
\end{figure}

To better understand how the model dynamically utilizes different patch lengths, Fig.~\ref{fig:heatmap_patchlen_irregular} shows a heatmap of the average attention weight assigned to each patch length across datasets and forecast ratios by \approach{}-R, averaged over multiple random seeds. The distribution varies significantly depending on the dataset and $\nicefrac{H}{M}$ value, indicating that \approach{}-R adaptively emphasizes different patch lengths based on task characteristics.

\subsection{Efficiency Analysis}\label{sec:effi_ana}

\subsubsection{Efficiency Improvement over ODE-based Models}
We compare the computational efficiency of \approach{} with several ODE-based models designed for irregular time series forecasting, including Latent-ODE~\cite{latentode:rubanova2019latent}, IVP-VAE~\cite{ivpvae:xiao2024ivp}, and ContiFormer~\cite{Contiformer:chen2023}. From the forecasting experiments in Section~\ref{sec:exp_forecasting}, we record the training time per epoch (T-epoch) and the number of epochs until early stopping (\# Epochs) to assess training efficiency. We also compute the per-sequence forward latency (T-forward) by dividing the inference time of a batch by its batch size, to evaluate inference efficiency. All measurements are obtained under the $H/M = 1.00$ setting and averaged across eight irregular time series datasets. To ensure a fair comparison, all experiments are conducted in the same computing environment equipped with NVIDIA A100 GPUs.

\begin{table}[ht]
	\footnotesize
	\centering
	\caption{Efficiency Comparison Across Models}
	\label{tab:efficiency}
	\setlength{\tabcolsep}{5pt}
	\begin{tabular}{lccc}
		\toprule
		Model       & T-epoch (s) & \# Epochs & T-forward (ms) \\
		\midrule
		Latent-ODE  & 292.9       & 38.4      & 916            \\
		IVP-VAE     & 11.2        & 68.0      & 12             \\
		ContiFormer & 119.4       & 43.7      & 217            \\
		FlexTSF     & 15.9        & 34.1      & 23             \\
		\bottomrule
	\end{tabular}
\end{table}

As shown in Table~\ref{tab:efficiency}, Latent-ODE incurs the highest computational cost due to its reliance on numerical integration over the entire sequence, resulting in prolonged training and inference times. IVP-VAE significantly reduces computation but requires more epochs to converge. FlexTSF is relatively efficient while maintaining the strong accuracy demonstrated in Section~\ref{sec:exp_forecasting}. Moreover, FlexTSF exhibits strong scalability and generalization in the pre-training and zero-shot experiments, where other ODE-based models are impractical due to their restricted model structures.

\subsubsection{Pre-trained Patcher and Convergence Efficiency}
We further examine how the IVP Patcher’s generalization ability accelerates convergence in the attention stack. From a Transformer perspective, both the IVP Patcher and linear patchers~\cite{PatchTST:nie2023time} act as tokenizers. Although FlexTSF can be trained end-to-end, we isolate the tokenizer’s effect by first pretraining both patcher types on the same dataset pool $\mathcal{D}_{p}$ under identical settings. For the linear patcher, we attach a linear decoder to form an autoencoder, enabling the same reconstruction objective as the IVP Patcher. We then freeze the pretrained tokenizer parameters and train the remaining components.
\begin{figure}[htbp]
	\centering
	\includegraphics[width=0.7\linewidth]{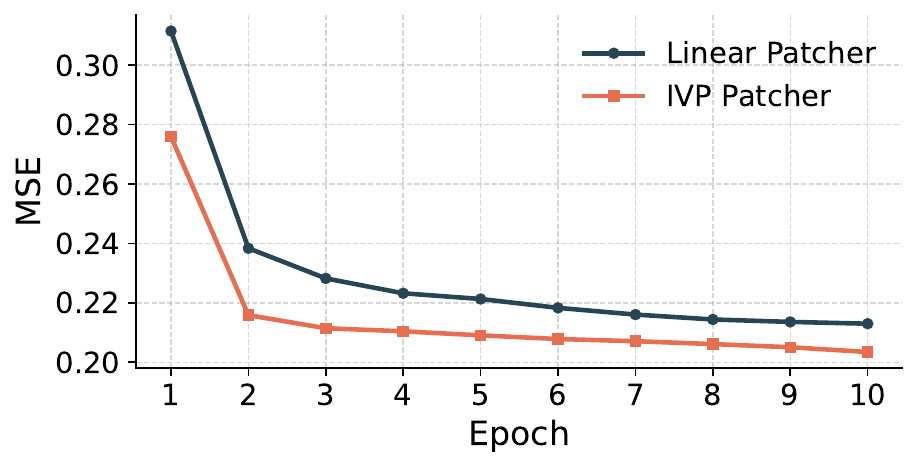}
	\caption{Training MSE of the model with frozen tokenizers. Initializing with the pretrained IVP Patcher yields consistently lower error and faster convergence than a linear patcher.}
	\label{fig:attn_mse_curve}
\end{figure}

Fig.~\ref{fig:attn_mse_curve} shows the MSE trend on the forecasting horizon over 10 training epochs. After just one epoch, the model with the IVP Patcher already achieves a lower MSE than its linear counterpart. These results indicate that, as a tokenizer, the pretrained IVP Patcher provides more informative representations that facilitate faster convergence of the attention layers.

\section{Conclusion}
We have presented \approach{}, a flexible forecasting model designed to handle time series data with diverse and irregular temporal characteristics. By introducing the IVP Patcher, \approach{} overcomes the limitations of fixed-window patching and learns continuous-time latent dynamics capable of handling arbitrary sampling intervals and missing values. Our time normalization strategy, combined with the Leader node, enables domain self-adaptation by decoupling static domain information extraction from temporal pattern learning. The use of random patch lengths further enhances \approach{}’s flexibility, reducing manual tuning and improving performance in modeling long-term dependencies.

A limitation of \approach{} and other single-modality time series models is that when input time series provide insufficient information about future regime changes or peak magnitudes, they cannot generate forecasts that distinguish among the possible futures. Incorporating exogenous signals or additional modalities into forecasting is an important direction for future work. Other directions include improving the efficiency of the IVP Patcher by exploring faster IVP solvers, extending the current channel-independent framework to fully multivariate inputs, developing adaptive patch length sampling strategies (e.g., via reinforcement learning), and pre-training on larger datasets to clarify the relationship between data scale and generalization.

\bibliographystyle{IEEEtran}
\bibliography{reference}

\begin{IEEEbiography}[{\includegraphics[width=1in,height=1.25in,clip,keepaspectratio]{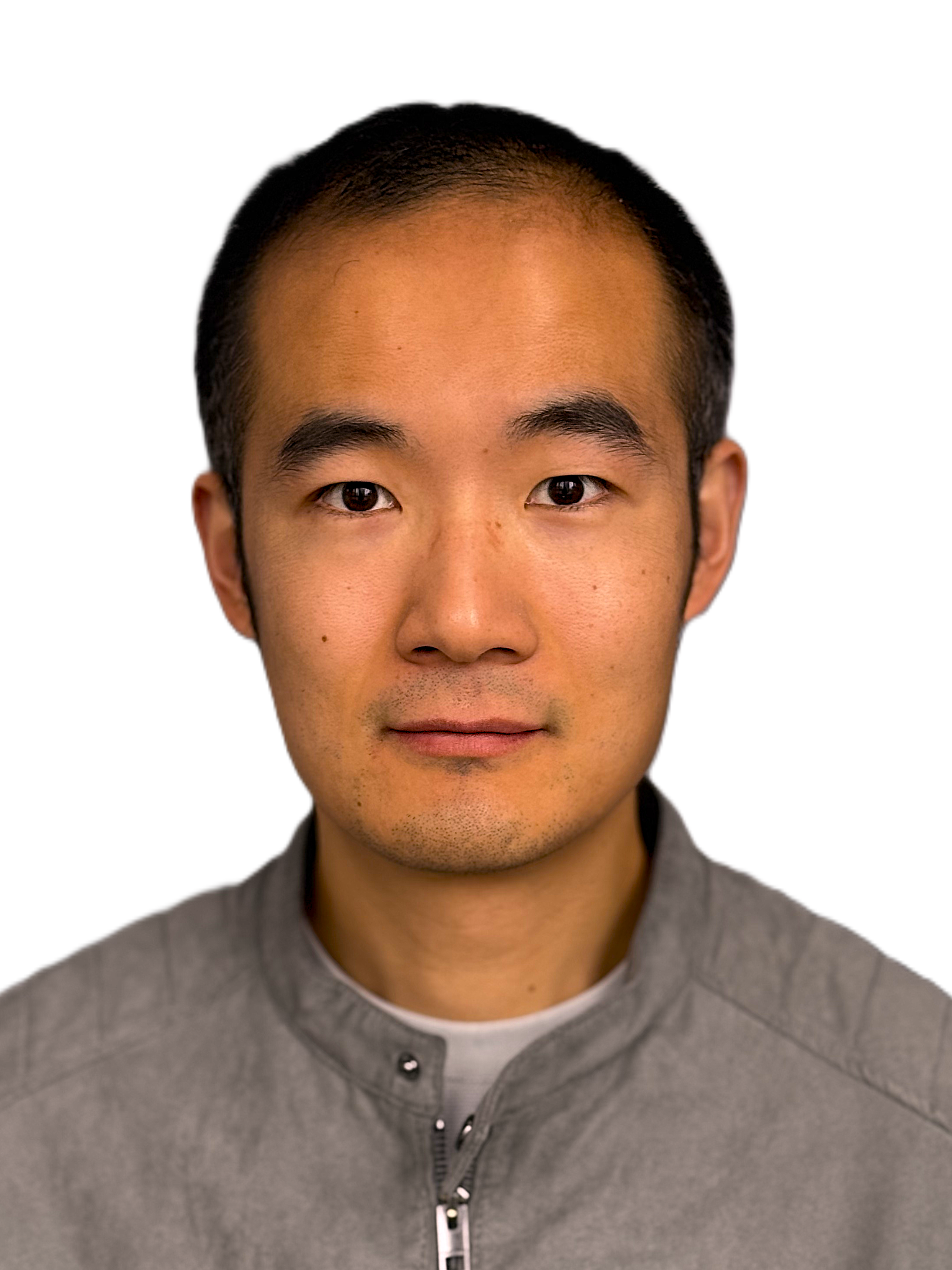}}]{Jingge Xiao}
	received the MSc degree from the University of Chinese Academy of Sciences. He is currently working toward the PhD degree at the L3S Research Center of the Leibniz Universität Hannover. Jingge’s ongoing research is focused on time series analysis and deep generative modeling.
\end{IEEEbiography}
\begin{IEEEbiography}
	[{\includegraphics[width=1in,height=1.25in,clip,keepaspectratio]{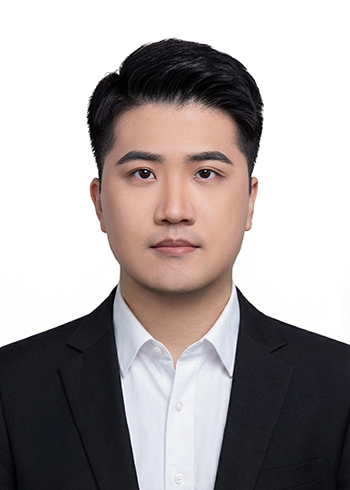}}]{Yile Chen}
	received the Ph.D. degree from Nanyang Technological University, Singapore, in 2022, and his B.Eng degree from Wuhan University, China, in 2018. He is currently an Associate Professor at the School of Computer Science and Engineering, South China University of Technology, China. His current research interests include data mining and artificial intelligence.
\end{IEEEbiography}
\begin{IEEEbiography}
	[{\includegraphics[width=1in,height=1.25in,clip,keepaspectratio]{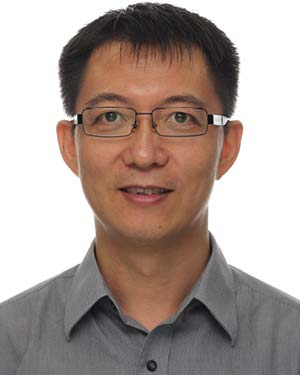}}]{Gao Cong}
	is a Professor and the Division Head of Data Science at the College of Computing and Data Science, Nanyang Technological University. Prior to joining NTU, he worked with Aalborg University, Microsoft Research Asia, and the University of Edinburgh. He received the Ph.D. degree from National University of Singapore in 2004. His current research interests include data management and spatio-temporal data mining.
\end{IEEEbiography}
\begin{IEEEbiography}
	[{\includegraphics[width=1in,height=1.25in,clip,keepaspectratio]{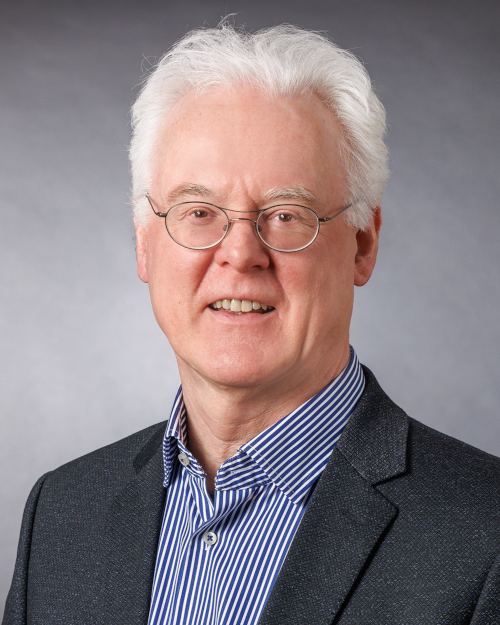}}]{Wolfgang Nejdl}
	was born in 1960. He received the Dipl.Ing. and Dr.Techn. degrees in computer science from Vienna University of Technology, in 1984 and 1988, respectively. He has been a Full Professor of computer science with Leibniz University Hannover, since 1995. He was an Assistant Professor in Vienna, from 1988 to 1992, and an Associate Professor at RWTH Aachen University, from 1992 to 1995. He was a Visiting Researcher/a Professor with Xerox PARC, Stanford University, the University of Illinois at Urbana-Champaign, EPFL Lausanne, PUC Rio, Trento, and Politecnico di Milano. He heads the L3S Research Center, www.L3S.de, and the Data Science Institute/Knowledge-Based Systems at Leibniz University. He has published more than 520 scientific articles listed at DBLP, with an H-index (based on Google Scholar) of 80. His research interests include information retrieval/search, web science, artificial intelligence, social and semantic web, digital libraries, and technology enhanced learning.
\end{IEEEbiography}
\begin{IEEEbiography}
	[{\includegraphics[width=1in,height=1.25in,clip,keepaspectratio]{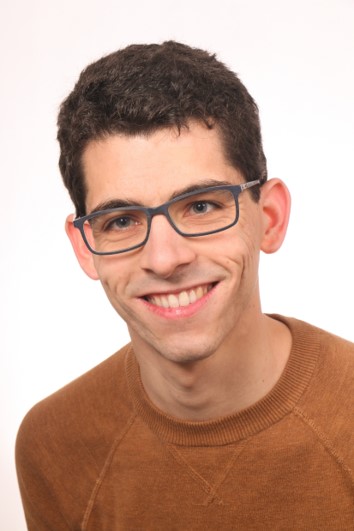}}]{Simon Gottschalk}
	has been a research group leader at the L3S Research Center of the Leibniz Universität Hannover since November 2022. Previously, he worked at the L3S Research Center since 2015 and completed his PhD there in April 2021 on the topic of “Creation, enrichment and application of knowledge graphs”. Research focuses in his group include knowledge graphs, spatio-temporal data analytics and Large Language Models. In different projects, he has conducted research on web archives, multilingual analysis of events, the development and application of event-based knowledge graphs and recommendations in mobility. His works have been cited more than 800 times and has an h-index of 13 according to Google Scholar.
\end{IEEEbiography}

\end{document}